\documentclass[10pt,twocolumn,letterpaper]{article}
\usepackage{cvpr}
\usepackage{cuted}
\def\confName{CVPR}
\def\confYear{2027}

\usepackage{amsmath,amsfonts,bm}

\def\eqref#1{equation~\ref{#1}}

\def\1{\bm{1}}

\DeclareMathAlphabet{\mathsfit}{\encodingdefault}{\sfdefault}{m}{sl}
\SetMathAlphabet{\mathsfit}{bold}{\encodingdefault}{\sfdefault}{bx}{n}

\usepackage{amsmath}
\usepackage{amssymb}
\usepackage{booktabs}
\usepackage{float}
\usepackage{placeins}
\usepackage{graphicx}

\usepackage{mathtools}
\usepackage{multirow}
\usepackage{subcaption}
\usepackage{url}
\usepackage{xcolor}
\usepackage{xspace}
\newcommand{\method}{InfiniHand\xspace}

\newcommand{\blfootnote}[1]{%
  \begingroup
  \renewcommand{\thefootnote}{}%
  \footnotetext[0]{#1}%
  \endgroup
}

\definecolor{cvprblue}{rgb}{0.21,0.49,0.74}
\usepackage[breaklinks,colorlinks,allcolors=cvprblue]{hyperref}
\title{InfiniHand: Streaming World-Space Hand Motion \\ Reconstruction from Egocentric Video}
\author{%
Kerui Ren$^{1,2}$ \quad
Kaiwen Song$^{1,3}$ \quad
Weiguang Zhao$^{1,4}$ \quad
Yuxi Wang$^{5}$ \quad
Yufei Liu$^{2}$ \\
Bo Dai$^{6}$ \quad
Haoyu Guo$^{1}$ \quad
Chunhua Shen$^{7,1}$ \quad
Mulin Yu$^{1\dagger}$ \quad
Tao Lu$^{1\dagger}$ \quad
Junting Dong$^{1}$ \\
{\small $^{1}$Shanghai Artificial Intelligence Laboratory, $^{2}$Shanghai Jiao Tong University,} \\
{\small $^{3}$University of Science and Technology of China, $^{4}$University of Liverpool,} \\
{\small $^{5}$Nanyang Technological University, $^{6}$The University of Hong Kong, $^{7}$Zhejiang University}%
}
\begin{document}
\maketitle

\begin{strip}
\centering
\vspace{-20pt}
\includegraphics[width=\textwidth]{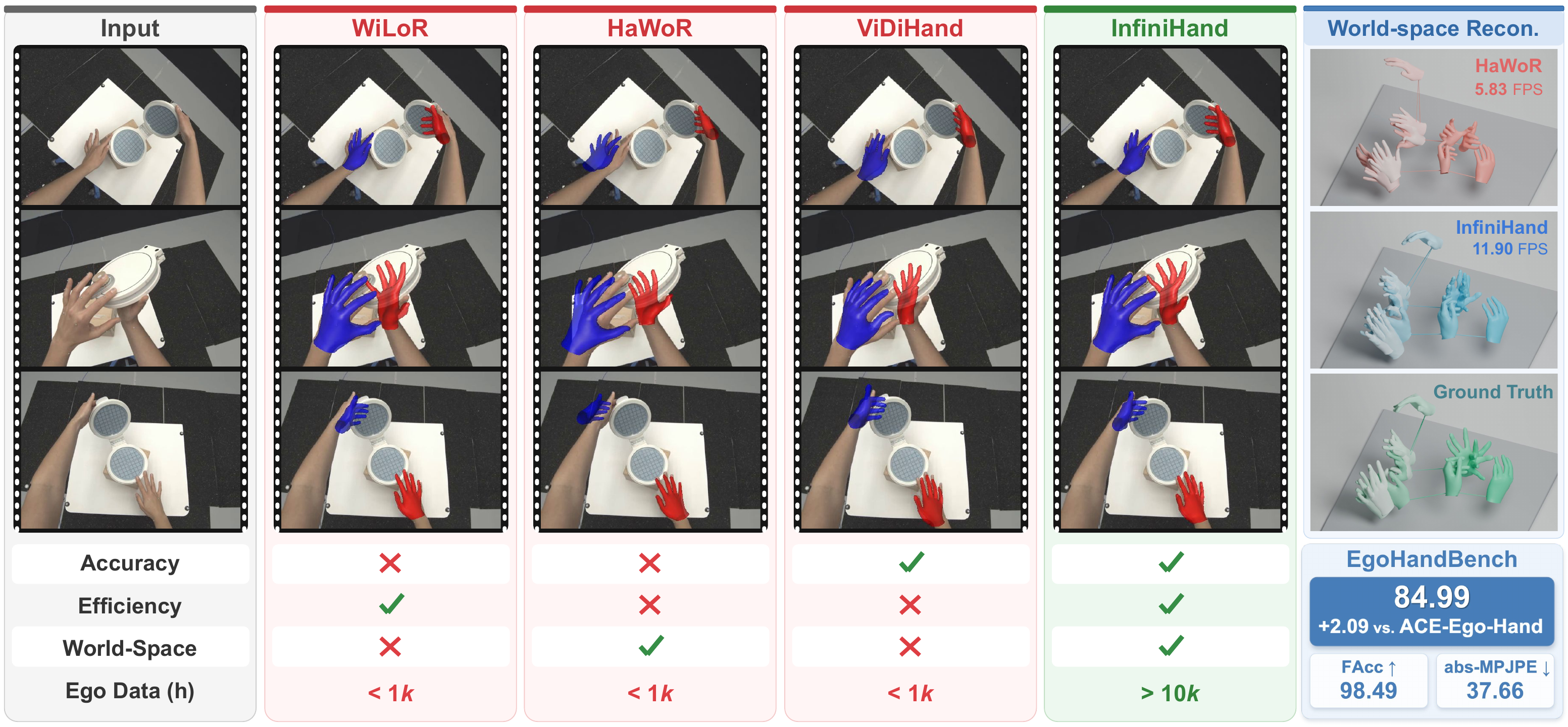}
\captionof{figure}{InfiniHand is a streaming framework for accurate and efficient world-space hand reconstruction, pretrained on over 10,000 hours of egocentric video. Under the unified EgoHandBench protocol, it achieves the best overall score and outperforms ACE-Ego-Hand. Project page: 
\href{https://infinihand.github.io/}{\textcolor{magenta}{\textbf{https://infinihand.github.io/}}}.}
\label{fig:teaser}
\end{strip}
\blfootnote{$^\dagger$Corresponding author.}
\begin{abstract}
World-space hand motion reconstruction from egocentric video requires recovering 3D articulated hand geometry while tracking camera egomotion. Existing approaches often cascade independent hand reconstruction modules with standalone camera tracking or geometric reconstruction systems, resulting in error accumulation, complex pipelines, and severe computational overhead. To address these limitations, we present InfiniHand, an end-to-end streaming feed-forward framework that reconstructs articulated hands while jointly predicting hand locations and camera trajectories directly from uncalibrated egocentric video. InfiniHand integrates persistent spatiotemporal memory with hand-centered visual features, explicitly coupling camera motion with local hand geometry within a unified architecture. We train InfiniHand in two progressive stages by first learning robust camera-space hand priors and then extending to streaming world-space reconstruction.
Extensive evaluations demonstrate that InfiniHand outperforms state-of-the-art baselines on in-domain benchmarks, reducing PA-MPJPE by 42.4\% on ARCTIC compared with ViDiHand while mitigating world-space drift. Furthermore, InfiniHand generalizes robustly to in-the-wild videos and operates at 11.90~FPS, delivering more than twice the throughput of HaWoR.
\end{abstract}

\section{Introduction}
\label{sec:introduction}

Egocentric video captures diverse human movements and complex hand–object interactions from the first-person perspective~\citep{grauman2022ego4d,hoque2025egodex}, where recovering hand motion in a shared world coordinate system transforms raw visual observations into actionable geometric demonstrations for embodied learning. These demonstrations enable training world action models via human-to-robot motion transfer~\citep{li2026egowam} and facilitate in-context robot imitation conditioned on retrieved human examples~\citep{papagiannis2025rx}. However, unconstrained in-the-wild videos rarely come paired with the ground-truth 3D hand poses and camera trajectories necessary to anchor movement within physical environments. Bridging this gap demands a scalable framework for rapid, high-fidelity world-space hand reconstruction, which is a critical prerequisite for downstream embodied learning.

Monocular hand reconstruction has advanced through transformer- and graph-based architectures~\citep{lin2021metro,lin2021meshgraphormer}, yet conventional world-space hand motion reconstruction pipelines still rely on separate models for hand localization, MANO~\citep{romero2017mano} reconstruction, and camera trajectory estimation. HaWoR~\citep{zhang2025hawor}, for example, combines hand detection and tracking, a dedicated camera-space hand reconstruction network, and DROID-SLAM~\citep{teed2021droidslam} with Metric3D~\citep{yin2023metric3d} for metric camera motion estimation. Coordinating these components introduces substantial computational and engineering overhead, while detection jitter, camera tracking drift, and scale inconsistencies can propagate through the pipeline and compromise world-space reconstruction. Recent feed-forward geometric models~\citep{leroy2024mast3r,wang2025vggt,wang2025cut3r,wang2025spann3r} provide powerful scene representations, but do not explicitly model articulated hand geometry. Beyond these architectural constraints, established reconstruction methods~\citep{potamias2024wilor, zhang2025hawor} lack extensive pretraining on diverse, unconstrained egocentric videos, rendering generalization to complex interactions and rapid camera motion a persistent bottleneck. Collectively, these drawbacks hinder accurate and efficient hand motion reconstruction from in-the-wild egocentric videos.

To address these limitations, we present \textbf{InfiniHand}, a streaming feed-forward framework that jointly predicts hand locations, MANO~\citep{romero2017mano} parameters, and camera trajectories within a unified architecture. Built upon a streaming 3D foundation model~\citep{chen2026lingbotmap}, InfiniHand establishes a shared spatiotemporal representation that couples global camera motion with local hand articulation. For each incoming frame, full-image geometric features guide hand mask prediction for precise localization, and the masked geometric features are subsequently fused with hand-centered appearance features to regress MANO parameters. In parallel, we fine-tune the camera head on egocentric sequences and integrate sparse keyframe bundle adjustment to ensure global trajectory consistency. To train this architecture effectively, our two-stage strategy first learns camera-space hand reconstruction and then jointly optimizes hand and camera motion over longer sequences. Together, these designs enable accurate, efficient camera-space and world-space hand reconstruction with strong generalization to challenging in-the-wild videos.

In summary, our primary contributions are as follows:
\begin{itemize}
    \item We introduce InfiniHand, an end-to-end framework built upon a streaming 3D foundation model that leverages a shared geometric representation to unify hand localization, MANO reconstruction, and camera motion prediction for world-space hand reconstruction.
    \item We adapt the camera head to egocentric dynamics via targeted fine-tuning and integrate sparse keyframe bundle adjustment to enforce global trajectory consistency.
    \item Extensive experiments demonstrate that InfiniHand achieves state-of-the-art accuracy in both camera- and world-space hand motion reconstruction with high streaming efficiency, while evaluations on in-the-wild videos confirm its strong generalization across complex real-world scenarios.
\end{itemize}

\section{Related Work}
\label{sec:related_work}
\paragraph{Hand Motion Reconstruction.}
Hand reconstruction has progressed from isolated mesh recovery to modeling bimanual interactions and global trajectories. For single-hand reconstruction, HaMeR~\citep{pavlakos2024hamer} scales up vision transformers, while WiLoR~\citep{potamias2024wilor} unifies hand localization and detailed mesh recovery in unconstrained scenes. Hamba~\citep{dong2024hamba} introduces graph-guided state-space modeling to capture joint spatial relations, and WildHands~\citep{prakash2024wildhands} adapts reconstruction to egocentric views. Although these approaches strengthen camera-space hand reconstruction, they do not recover camera egomotion or world-space trajectories.

Extending to bimanual and temporal modeling, InterWild~\citep{moon2023interwild} decouples per-hand recovery from relative translation, while OmniHands~\citep{lin2024omnihands} leverages spatiotemporal reasoning for interacting hands. More recently, ViDiHand~\citep{wang2026vidihand} adapts video diffusion priors with hand-overlay supervision for temporally coherent egocentric geometry, and ACE-Ego-Hand~\citep{liu2026ace} repurposes video diffusion into a deterministic latent encoder with bidirectional decoding to recover metric bimanual trajectories across severe occlusions. Nevertheless, recovering world-space motion requires disentangling hand movement from camera egomotion. To this end, HaWoR~\citep{zhang2025hawor} combines egocentric SLAM with motion infilling, Dyn-HaMR~\citep{yu2025dynhamr} applies multi-stage optimization over camera tracking and interacting-hand priors, and MINT~\citep{zhu2026mint} jointly predicts camera trajectories, camera-frame hand states, and hand presence from a shared spatiotemporal representation to obtain world-space motion via explicit coordinate transformations.

\paragraph{Streaming 3D Reconstruction.}
Reconstructing scene geometry and camera motion from video has traditionally relied on simultaneous localization and mapping, where visual tracking is coupled with bundle adjustment. Classic frameworks such as ORB-SLAM2~\citep{murartal2017orbslam2} pair sparse feature tracking with keyframe loop closure, whereas DROID-SLAM~\citep{teed2021droidslam} introduces learned recurrent updates and differentiable dense bundle adjustment. Building on this paradigm, learning-augmented SLAM systems incorporate feed-forward geometric priors into optimization. Specifically, MASt3R-SLAM~\citep{murai2025mast3rslam} anchors tracking and global optimization on two-view 3D reconstruction models, VGGT-SLAM~\citep{maggio2025vggtslam} aligns feed-forward submaps via projective optimization and loop closure, and M$^3$~\citep{ren2026m3} equips multi-view foundation models with dense matching heads for monocular Gaussian splatting SLAM. Although these hybrid pipelines benefit from learned priors, they still depend on explicit optimization for cross-view consistency. In contrast, purely feed-forward architectures maintain geometric context natively within the network without post-hoc optimization. For example, LoGeR~\citep{zhang2026loger} combines sliding-window attention with test-time-training memory across video chunks to preserve local detail and long-range consistency, while LingBot-Map~\citep{chen2026lingbotmap} employs a streaming context transformer with anchor context, a pose-reference window, and trajectory memory for incremental reconstruction over extended sequences.

\section{Method}
\label{sec:method}

\begin{figure*}[t]
\centering
\includegraphics[width=\textwidth]{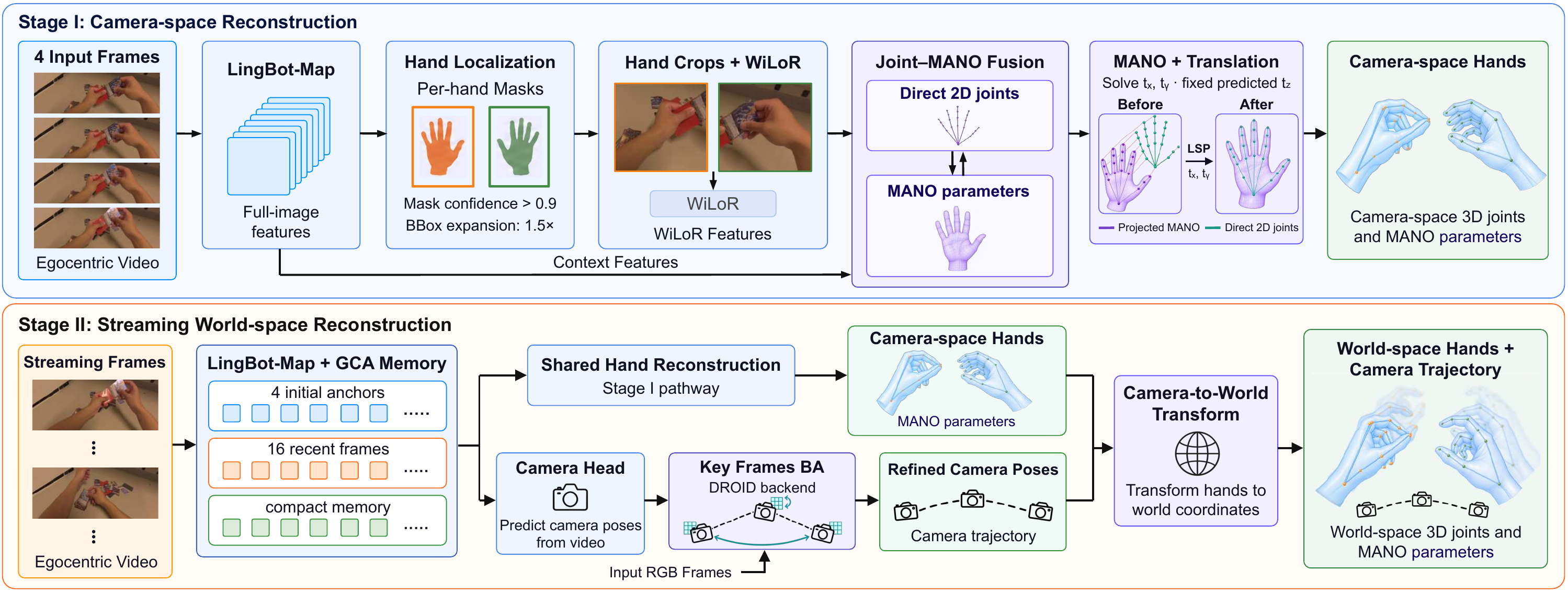}
\vspace{-14pt}
\caption{\textbf{Overview of InfiniHand.} Stage I learns hand localization and MANO reconstruction from geometric and appearance features, transforming hand-frame predictions into camera coordinates. Stage II jointly reconstructs hands and estimates camera trajectories with streaming memory, followed by sparse bundle adjustment for camera refinement and world-space reconstruction.}
\label{fig:method_pipeline}
\end{figure*}

Fig.~\ref{fig:method_pipeline} presents the overall pipeline of \method, a streaming framework for articulated hand reconstruction and camera parameter estimation from egocentric video. Given an input RGB sequence $\mathcal I=\{\mathbf I_t\}_{t=1}^{T}$, we use $t\in\{1,\ldots,T\}$ to index frames and $s\in\{\mathrm L,\mathrm R\}$ to index the hand side. For each frame $t$, the model predicts camera parameters $\hat P_t=(\hat{\mathbf K}_t,\hat{\mathbf R}_t,\hat{\mathbf u}_t)$, where $\hat{\mathbf K}_t$ denotes the intrinsic matrix and $(\hat{\mathbf R}_t,\hat{\mathbf u}_t)$ represents the camera-to-world pose, yielding the complete camera sequence $\hat{\mathcal P}=\{\hat P_t\}_{t=1}^{T}$. Superscripts $\mathrm h$, $\mathrm c$, and $\mathrm w$ denote the virtual hand-camera, original camera, and first-anchor world coordinate systems, respectively. For each hand $s$, the final articulated state comprises MANO~\citep{romero2017mano} pose $\hat{\boldsymbol{\Theta}}_t^s\in\mathbb R^{15\times3}$, shape $\hat{\boldsymbol{\beta}}_t^s\in\mathbb R^{10}$, global orientation $\hat{\boldsymbol{\Phi}}_t^{\mathrm{w},s}\in\mathbb R^3$, and wrist translation $\hat{\mathbf t}_t^{\mathrm{w},s}\in\mathbb R^3$. The model first predicts hand-frame geometry, converts it to camera coordinates, and then obtains world-space motion using the estimated camera trajectory. Specifically, Sec.~\ref{sec:data_preprocessing} describes data preprocessing, Sec.~\ref{sec:camera_space} introduces camera-space hand localization and reconstruction, and Sec.~\ref{sec:world_space} presents streaming world-space reconstruction and sparse geometric refinement.

\subsection{Training Data Preprocessing}
\label{sec:data_preprocessing}

We train InfiniHand using existing annotations from seven public hand motion reconstruction datasets: ARCTIC~\citep{fan2023arctic}, DexYCB~\citep{chao2021dexycb}, EgoSuite~\citep{egostandard}, EgoVerse~\citep{punamiya2026egoverse}, HOT3D~\citep{banerjee2025hot3d}, H2O3D~\citep{hampali2022keypointtransformer}, and HO3D~\citep{hampali2020honnotate}. Video frames are uniformly sampled at 10 FPS. We use the provided MANO parameters when available; otherwise, we fit MANO to the annotated 3D hand keypoints and temporally align the fitted parameters with the sampled frames. Handedness and coordinate conventions are standardized across datasets. We decode 3D joint targets from the corresponding MANO parameters and project them using the camera intrinsics to obtain consistent 2D supervision.

For hand localization, we render the left- and right-hand MANO meshes into each image to generate per-hand segmentation masks, from which tight bounding boxes are derived. Each sample comprises an RGB image, camera-space MANO parameters and 3D joints, per-hand masks, and bounding boxes. For sequences with ground-truth camera trajectories, we retain camera pose annotations and transform camera-space hand parameters and joints into a shared world coordinate frame to supervise joint hand and camera trajectory reconstruction. Training data are assigned to the two stages according to supervision availability: Stage I uses all valid camera-space hand annotations, while Stage II additionally uses the subset with camera trajectories for camera pose and world-space joint supervision.

\subsection{Camera-Space Hand Reconstruction}
\label{sec:camera_space}

\paragraph{Hand localization.}
Our hand localization module leverages full-image geometric features from the pretrained LingBot-Map backbone~\citep{chen2026lingbotmap} to identify left- and right-hand regions. Let $\mathbf F_t$ denote the full-image geometric representation for frame $t$. A Dense Prediction Transformer (DPT)~\citep{ranftl2021dpt} decoder aggregates four intermediate backbone layers into a dense full-image mask feature $\mathbf F_{t,\mathrm{mask}}$. We initialize the decoder from a pretrained depth head, retaining its feature projections, multi-scale resizing, and coarse-to-fine RefineNet~\citep{lin2017refinenet} fusion, while replacing the final depth prediction layer with a randomly initialized two-channel mask head. The localization module predicts
\begin{equation}
[\hat{\mathbf m}_t^{\mathrm L},\hat{\mathbf m}_t^{\mathrm R}]
= \sigma\!\left(\mathcal H_{\mathrm{mask}}(\mathbf F_{t,\mathrm{mask}})\right),
\end{equation}
where $\hat{\mathbf m}_t^s$ is the predicted mask probability map for hand $s$, $\mathcal H_{\mathrm{mask}}$ upsamples the decoded feature to the input image resolution, and $\sigma(\cdot)$ denotes the element-wise sigmoid function. With frame and hand indices omitted when aggregating the loss, the localization objective combines binary cross-entropy and Dice terms:
\begin{equation}
\mathcal L_{\mathrm{mask}} =
\lambda_{\mathrm{BCE}}\mathcal L_{\mathrm{BCE}}(\hat{\mathbf m},\mathbf m)
+ \lambda_{\mathrm{Dice}}\mathcal L_{\mathrm{Dice}}(\hat{\mathbf m},\mathbf m),
\end{equation}
where $\mathbf m_t^s$ is the corresponding binary supervision. Validity flags ignore missing annotations to avoid false negative supervision. Subsequently, thresholded masks are converted into expanded bounding boxes $\mathbf{b}_t^s$ for hand reconstruction.

\paragraph{Hand reconstruction.}
For each hand box $\mathbf b_t^s$, we crop the full-image geometric representation $\mathbf F_t$ and re-embed it into a hand-centered patch grid, yielding $\mathbf f_{t,\mathrm{geo}}^s$. In parallel, a WiLoR encoder~\citep{potamias2024wilor} processes the corresponding RGB crop to produce $\mathbf f_{t,\mathrm{rgb}}^s$. Both branches share the same cropping conventions, including horizontal flipping for left-hand canonicalization. We fuse the cropped geometric and RGB features with a $1\times1$ convolution:
\begin{equation}
\begin{gathered}
\mathbf f_{t,\mathrm{rgb}}^s
= \mathcal E_{\mathrm{WiLoR}}(\operatorname{Crop}(\mathbf I_t,\mathbf b_t^s)),\\
\mathbf f_t^s
= \operatorname{Conv}_{1\times1}([\mathbf f_{t,\mathrm{geo}}^s;\mathbf f_{t,\mathrm{rgb}}^s]).
\end{gathered}
\end{equation}

Driven by the fused hand feature $\mathbf f_t^s$, the MANO head jointly predicts normalized 2D hand joints, MANO parameters, and wrist depth in the virtual hand-camera frame $\mathrm h$. For readability, we omit the frame index $t$ and hand side $s$ in the following hand-space derivation:
\begin{equation}
\begin{gathered}
(\hat{\mathbf p},\hat{\boldsymbol{\Theta}},\hat{\boldsymbol{\beta}},\hat{\boldsymbol{\Phi}}^{\mathrm{h}},\hat\ell_{\mathrm{z}}^{\mathrm h}) = \mathcal H_{\mathrm{MANO}}(\mathbf f), \
\hat t_{\mathrm{z}}^{\mathrm{h}} = \exp(\hat\ell_{\mathrm{z}}^{\mathrm h}).
\end{gathered}
\end{equation}
Here, $\hat{\mathbf p}$ is expressed in normalized crop coordinates, and $\hat{\mathbf p}^{\mathrm{pix}}$ denotes the same joints after conversion to crop-pixel coordinates.

After reversing left-hand canonicalization, the MANO decoder~\citep{romero2017mano} maps the predicted pose, shape, and orientation into wrist-centered 3D joints $\bar{\mathbf J}^{\mathrm{h}}$. To recover the remaining lateral wrist translation $(\hat t_{\mathrm{x}}^{\mathrm{h}},\hat t_{\mathrm{y}}^{\mathrm{h}})$ within the hand frame, we follow ViDiHand~\citep{wang2026vidihand} by aligning the projected 3D joints with the predicted 2D joints $\hat{\mathbf p}^{\mathrm{pix}}$ and wrist depth $\hat t_{\mathrm{z}}^{\mathrm{h}}$ via differentiable least-squares projection (LSP), which solves for lateral translation while keeping the predicted depth fixed:
\begin{equation}
\hat{\mathbf t}^{\mathrm h}
= \underset{\mathbf t:\,t_{\mathrm z}=\hat t_{\mathrm z}^{\mathrm h}}{\arg\min}
\sum_{j\in\mathcal V}
\left\|\pi\!\left(\bar{\mathbf J}_{j}^{\mathrm{h}}+\mathbf t;\mathbf K^{\mathrm{h}}\right)-\hat{\mathbf p}_{j}^{\mathrm{pix}}\right\|_2^2,
\label{eq:lsp_translation}
\end{equation}
where $\mathbf K^{\mathrm{h}}$ is the virtual crop-camera intrinsic matrix induced by the full-image intrinsics and perspective crop, $\pi(\cdot;\mathbf K^{\mathrm{h}})$ is the perspective projection, $\hat{\mathbf p}_{j}^{\mathrm{pix}}$ are the predicted 2D joints in crop pixels, and $\mathcal V$ represents valid joints with positive depth.

With the recovered translation, we obtain the complete hand-frame MANO representation and transform its global orientation and translation into the original camera coordinate system:
\begin{equation}
\operatorname{Rot}(\hat{\boldsymbol{\Phi}}^{\mathrm{c}})=\mathbf R_{\mathrm{h}\leftarrow\mathrm{c}}^{\top}\operatorname{Rot}(\hat{\boldsymbol{\Phi}}^{\mathrm{h}}),\ 
\hat{\mathbf t}^{\mathrm{c}}=\mathbf R_{\mathrm{h}\leftarrow\mathrm{c}}^{\top}\hat{\mathbf t}^{\mathrm{h}}.
\label{eq:hand_to_camera}
\end{equation}
Here $\mathbf R_{\mathrm{h}\leftarrow\mathrm{c}}$ rotates from the original camera to the virtual hand camera, and $\operatorname{Rot}(\cdot)$ converts an axis-angle vector to its rotation matrix via the Rodrigues formula.

The MANO head is jointly supervised by 2D joints, pose, shape, global orientation, translation, 3D joints, and reprojection. The 2D joint term $\mathcal L_{\mathrm{2D}}$ is a mean $L_1$ loss over valid joints in normalized crop coordinates. Global orientation, translation, and 3D joints are compared in the original camera frame $\mathrm c$, while local pose and shape are frame-invariant and reprojection uses the corresponding image coordinates:
\begin{equation}
\begin{aligned}
\mathcal L_{\mathrm{MANO}} &=
\lambda_{\mathrm{2D}}\mathcal L_{\mathrm{2D}}+
\lambda_{\mathrm{o}}\mathcal L_{\mathrm{orient}}+
\lambda_{\mathrm{r}}\mathcal L_{\mathrm{reproj}} \\
+&\lambda_{\mathrm{s}}\mathcal L_{\mathrm{shape}}+
\lambda_{\mathrm{t}}\mathcal L_{\mathrm{trans}}+
\lambda_{\mathrm{j}}\mathcal L_{\mathrm{joints}}^{\mathrm{c}}+\lambda_{\mathrm{p}}\mathcal L_{\mathrm{pose}}
.
\end{aligned}
\end{equation}
In Stage I, the mask and MANO heads are optimized separately using the available camera-space supervision. Detailed formulations of all loss terms are provided in the supplementary materials.

\subsection{World-Space Hand Reconstruction}
\label{sec:world_space}

\paragraph{Joint streaming training.}
In Stage II, we integrate the pretrained mask and MANO heads with the native LingBot-Map camera decoder and jointly optimize them over streaming video clips initialized with anchor frames and partitioned into consecutive windows. For window $k$ with frame indices $\Omega_k$, let $\mathbf I_k=\operatorname{stack}_{t\in\Omega_k}\mathbf I_t$ and $\mathbf F_k=\operatorname{stack}_{t\in\Omega_k}\mathbf F_t$ denote the input frames and updated geometric features. Cross-window context is maintained by a streaming memory state $\mathcal M_k=(\mathcal M_{\mathrm{anchor}},\mathcal W_k,\mathcal T_k)$ managed via Geometric Context Attention (GCA)~\citep{chen2026lingbotmap}, which combines anchor features $\mathcal M_{\mathrm{anchor}}$ for a shared reference frame, a dense window $\mathcal W_k$ for fine-grained local correspondences, and compressed trajectory tokens $\mathcal T_k$ for historical context after dense features are evicted. This state persists across adjacent windows within a clip, resets between independent clips, and updates at window $k$ via $(\mathbf F_k,\mathcal M_k)=\mathcal B_{\mathrm{GCA}}(\mathbf I_k,\mathcal M_{k-1})$.

To decode frame-level geometry from the updated window features $\mathbf F_k$, each per-frame representation $\mathbf F_t$ ($t\in\Omega_k$) drives two parallel branches, where the camera decoder predicts $\hat P_t=(\hat{\mathbf K}_t,\hat{\mathbf R}_t,\hat{\mathbf u}_t)$ and the mask and MANO heads regress hand predictions as
\begin{equation}
\begin{aligned}
[\hat{\mathbf m}_t^{\mathrm L},\hat{\mathbf m}_t^{\mathrm R}]
=\sigma\!\left(\mathcal H_{\mathrm{mask}}(\mathbf F_{t,\mathrm{mask}})\right),\\
(\hat{\mathbf p}_t^s,\hat{\boldsymbol{\Theta}}_t^s,\hat{\boldsymbol{\beta}}_t^s,
\hat{\boldsymbol{\Phi}}_t^{\mathrm{h},s},\hat\ell_{\mathrm{z},t}^{\mathrm h,s})
=\mathcal H_{\mathrm{MANO}}(\mathbf f_t^s).
\end{aligned}
\end{equation}
Following Sec.~\ref{sec:camera_space}, LSP recovers the hand-centric translation $\hat{\mathbf t}_t^{\mathrm{h},s}$ from the predicted 2D joints, MANO joints, and wrist depth $\hat t_{\mathrm{z},t}^{\mathrm{h},s}=\exp(\hat\ell_{\mathrm{z},t}^{\mathrm h,s})$, after which $\hat{\boldsymbol{\Phi}}_t^{\mathrm{h},s}$ and $\hat{\mathbf t}_t^{\mathrm{h},s}$ are converted to camera coordinates to yield $\hat Y_t^s=(\hat{\mathbf p}_t^s,\hat{\boldsymbol{\Theta}}_t^s,\hat{\boldsymbol{\beta}}_t^s,\hat{\boldsymbol{\Phi}}_t^{\mathrm{c},s},\hat{\mathbf t}_t^{\mathrm{c},s})$. Combining $\hat Y_t^s$ with $\hat P_t$ yields the first-anchor world-frame orientation and translation via
\begin{equation}
\operatorname{Rot}(\hat{\boldsymbol{\Phi}}_t^{\mathrm{w},s})=\hat{\mathbf R}_t\operatorname{Rot}(\hat{\boldsymbol{\Phi}}_t^{\mathrm{c},s}),\quad
\hat{\mathbf t}_t^{\mathrm{w},s}=\hat{\mathbf R}_t\hat{\mathbf t}_t^{\mathrm{c},s}+\hat{\mathbf u}_t,
\label{eq:camera_to_world}
\end{equation}
while local articulated pose $\hat{\boldsymbol{\Theta}}_t^s$ and shape $\hat{\boldsymbol{\beta}}_t^s$ remain invariant under rigid transformation. During training, predicted and ground-truth 3D joints and translations undergo the same first-anchor transformation and sample-level normalization $\tilde{\mathbf x}=\mathbf x/\kappa$ with $\kappa>0$, whereas rotations and camera intrinsics remain unscaled.

For training losses, we combine mask supervision, camera-space MANO supervision including its 2D joint term, world-space joint supervision, temporal consistency, and camera supervision:
\begin{equation}
\begin{aligned}
\mathcal L_{\mathrm{StageII}}={}&\lambda_{\mathrm{c}}\mathcal L_{\mathrm{cam}}
+\lambda_{\mathrm{w}}\mathcal L_{\mathrm{joints}}^{\mathrm{w}}\\
+&\lambda_{\mathrm{temp}}\mathcal L_{\mathrm{temp}}
+\mathcal L_{\mathrm{MANO}}
+\mathcal L_{\mathrm{mask}}.
\label{eq:stage2_loss}
\end{aligned}
\end{equation}
Here, $\mathcal L_{\mathrm{MANO}}$ retains the camera-space supervision defined in Sec.~\ref{sec:camera_space}, including its 2D joint term, while $\mathcal L_{\mathrm{joints}}^{\mathrm{w}}$ constrains the reconstructed joints in the shared world frame. Specifically, the camera loss $\mathcal L_{\mathrm{cam}}$ combines absolute pose and field-of-view supervision with relative-motion supervision between valid frame pairs.  Meanwhile, $\mathcal L_{\mathrm{temp}}$ enforces temporal consistency across camera predictions, MANO parameters, hand masks, and 2D joints. Each loss term is evaluated conditionally based on annotation availability.

\paragraph{Sparse bundle adjustment.}
Although trajectory memory preserves long-range context, compressing dense features beyond the anchor and recent windows discards fine-grained geometric constraints from earlier observations, causing residual drift over extended sequences. To address this, we integrate a DROID bundle-adjustment backend~\citep{teed2021droidslam} into streaming prediction to enforce explicit cross-frame constraints. Rather than optimizing over the entire history, we maintain a binary keyframe pool that balances local temporal continuity with long-range geometric constraints, enabling sparse refinement to revisit past observations and curb cumulative camera drift. The refined camera poses then transform local hand predictions into a unified world coordinate system, with implementation details and pool update rules provided in the supplementary material.
\begin{table*}[t]
\centering
\caption{\textbf{Camera-space quantitative comparison.} Evaluating detection and camera-space hand motion metrics across four benchmarks, InfiniHand consistently yields the lowest PA. Abs-MPJPE, MPJPE, PA, and EPE$_{\mathrm{2D}}$ include missed-hand penalties. \textbf{Bold} and \underline{underlined} denote best and second-best results.}
\label{tab:camera_main}
\begingroup
\setlength{\tabcolsep}{2.2pt}
\renewcommand{\arraystretch}{1.08}
\fontsize{7}{8}\selectfont
\resizebox{\textwidth}{!}{%
\begin{tabular}{@{}l*{7}{c}@{\hspace{9pt}}*{7}{c}@{}}
\toprule
Method & \multicolumn{7}{c}{\textbf{ARCTIC}} & \multicolumn{7}{c}{\textbf{HOT3D}} \\
\cmidrule(lr){2-8}\cmidrule(lr){9-15}
& F1$\uparrow$ & FAcc$\uparrow$ & Abs-MPJPE$\downarrow$ & MPJPE$\downarrow$ & PA$\downarrow$ & EPE$_{\mathrm{2D}}\downarrow$ & Jitter$\downarrow$ & F1$\uparrow$ & FAcc$\uparrow$ & Abs-MPJPE$\downarrow$ & MPJPE$\downarrow$ & PA$\downarrow$ & EPE$_{\mathrm{2D}}\downarrow$ & Jitter$\downarrow$ \\
\midrule
InterWild & 0.801 & 0.647 & 108.11 & 44.80 & 17.60 & 50.56 & 66.10 & 0.724 & 0.435 & 175.63 & 81.90 & 18.51 & 48.57 & 81.44 \\
HaMeR & 0.940 & 0.846 & 65.41 & 30.81 & 13.24 & 26.74 & 16.44 & 0.884 & 0.698 & 101.66 & 66.26 & 13.38 & 23.75 & 21.23 \\
Hamba & 0.871 & 0.716 & 80.73 & 37.29 & 16.34 & 43.80 & 13.92 & 0.713 & 0.370 & 175.66 & 80.74 & 22.30 & 61.50 & 14.09 \\
WildHands & 0.935 & 0.836 & 56.95 & 28.01 & 13.35 & 32.31 & 12.52 & 0.771 & 0.583 & 177.65 & 59.44 & 20.28 & 69.17 & 22.31 \\
OmniHands & 0.882 & 0.754 & 88.51 & 32.69 & 12.26 & 27.78 & 54.01 & 0.839 & 0.566 & 125.66 & 61.39 & 12.60 & 22.88 & 60.90 \\
WiLoR & 0.967 & 0.902 & \underline{49.10} & 21.22 & 9.96 & 18.25 & 23.08 & 0.948 & 0.851 & 74.06 & 31.32 & 10.49 & 17.26 & 18.11 \\
Dyn-HaMR & 0.977 & 0.942 & 194.55 & 29.30 & 11.33 & 15.66 & 24.37 & 0.858 & 0.619 & 97.23 & 34.18 & 13.98 & 18.72 & 14.80 \\
HaWoR & 0.953 & 0.889 & 57.95 & 25.27 & 10.14 & 28.65 & 16.95 & 0.950 & 0.868 & 68.29 & 27.53 & 9.58 & 16.10 & 15.46 \\
ViDiHand & 0.992 & 0.972 & 89.26 & 25.80 & 10.66 & 17.40 & 4.52 & \underline{0.981} & \underline{0.942} & \underline{49.96} & \underline{22.86} & 9.83 & \underline{10.40} & 4.61 \\
MINT & 0.970 & 0.911 & 89.89 & 30.70 & 11.71 & 31.54 & 13.46 & 0.974 & 0.921 & 65.60 & 23.88 & 9.80 & 16.12 & 11.90 \\
ACE-Ego-Hand & \underline{0.995} & \underline{0.982} & 75.76 & \underline{15.27} & \underline{7.71} & \underline{13.22} & \underline{2.67} & 0.974 & 0.921 & 130.59 & 24.63 & \underline{8.78} & 17.75 & \underline{2.85} \\
\midrule
Ours & \textbf{0.999} & \textbf{0.997} & \textbf{32.96} & \textbf{15.12} & \textbf{6.14} & \textbf{7.65} & \textbf{2.35} & \textbf{0.989} & \textbf{0.964} & \textbf{45.06} & \textbf{19.88} & \textbf{7.69} & \textbf{7.63} & \textbf{2.31} \\
\bottomrule
\end{tabular}}
\par\vspace{2pt}
\resizebox{\textwidth}{!}{%
\begin{tabular}{@{}l*{7}{c}@{\hspace{9pt}}*{7}{c}@{}}
\toprule
Method & \multicolumn{7}{c}{\textbf{HOI4D}} & \multicolumn{7}{c}{\textbf{Gen-HumanEgo}} \\
\cmidrule(lr){2-8}\cmidrule(lr){9-15}
& F1$\uparrow$ & FAcc$\uparrow$ & Abs-MPJPE$\downarrow$ & MPJPE$\downarrow$ & PA$\downarrow$ & EPE$_{\mathrm{2D}}\downarrow$ & Jitter$\downarrow$ & F1$\uparrow$ & FAcc$\uparrow$ & Abs-MPJPE$\downarrow$ & MPJPE$\downarrow$ & PA$\downarrow$ & EPE$_{\mathrm{2D}}\downarrow$ & Jitter$\downarrow$ \\
\midrule
InterWild & 0.530 & 0.039 & 208.71 & 59.55 & 17.38 & 55.34 & 195.28 & 0.893 & 0.827 & 92.28 & 61.24 & 13.26 & 25.02 & 49.62 \\
HaMeR & 0.858 & 0.730 & 104.27 & 40.70 & 13.83 & 28.60 & 24.30 & 0.939 & 0.897 & 57.64 & 47.14 & 10.87 & 14.87 & 14.45 \\
Hamba & 0.683 & 0.255 & 100.24 & 41.24 & 13.59 & 25.06 & 19.31 & 0.873 & 0.721 & 87.29 & 54.28 & 14.85 & 28.08 & 14.46 \\
WildHands & 0.842 & 0.699 & 148.09 & 47.74 & 16.76 & 40.61 & 18.85 & 0.445 & 0.239 & 240.44 & 77.36 & 31.47 & 102.86 & 17.98 \\
OmniHands & 0.625 & 0.156 & 98.21 & 39.68 & 11.41 & 17.38 & 39.51 & 0.940 & 0.905 & 70.46 & 45.42 & 10.85 & 15.99 & 19.41 \\
WiLoR & 0.965 & 0.953 & 68.12 & 32.92 & 11.66 & 16.41 & 16.08 & 0.966 & 0.940 & \underline{42.58} & 40.49 & 8.93 & 10.87 & 9.43 \\
Dyn-HaMR & 0.940 & 0.918 & 127.87 & 31.08 & 12.73 & 16.18 & 11.72 & 0.970 & 0.949 & 51.79 & \underline{21.08} & 10.27 & \underline{8.46} & 8.43 \\
HaWoR & 0.915 & 0.869 & 62.72 & 29.45 & 11.98 & 14.77 & 21.92 & 0.968 & 0.948 & 43.80 & 21.12 & 9.60 & 9.60 & 13.42 \\
ViDiHand & 0.970 & 0.956 & 93.19 & 25.54 & \underline{9.69} & 20.66 & 3.98 & 0.941 & 0.859 & 267.47 & 38.20 & 12.67 & 22.00 & 5.32 \\
MINT & 0.960 & 0.937 & 84.93 & 30.87 & 12.21 & 21.68 & 13.59 & \underline{0.973} & \underline{0.963} & 43.28 & 22.05 & \underline{8.32} & 9.50 & 8.41 \\
ACE-Ego-Hand & \underline{0.980} & \underline{0.971} & \underline{54.47} & \underline{22.88} & 11.30 & \underline{11.70} & \underline{2.44} & 0.963 & 0.931 & 149.65 & 24.30 & 9.61 & 14.13 & \underline{2.37} \\
\midrule
Ours & \textbf{0.988} & \textbf{0.981} & \textbf{50.98} & \textbf{18.86} & \textbf{7.71} & \textbf{8.18} & \textbf{1.89} & \textbf{0.989} & \textbf{0.976} & \textbf{40.55} & \textbf{19.52} & \textbf{7.92} & \textbf{6.90} & \textbf{1.81} \\
\bottomrule
\end{tabular}}
\endgroup
\vspace{-8pt}
\end{table*}

\section{Experiments}
\label{sec:experiments}

\subsection{Experimental Setup}
\label{sec:experimental_setup}
\paragraph{Datasets \& Metrics.}
For camera-space evaluation, we assess in-domain performance on 34 test scenes from ARCTIC~\citep{fan2023arctic} and 10 from HOT3D~\citep{banerjee2025hot3d}, while evaluating out-of-domain generalization on 166 test scenes from HOI4D~\citep{liu2022hoi4d} and 100 from Gen-HumanEgo~\citep{genrobot2026genhumanego}. 
Among these benchmarks, we further conduct world-space evaluation on ARCTIC, HOT3D, and Gen-HumanEgo.
We evaluate hand detection using Frame Accuracy (FAcc) and F1 score. Camera-space reconstruction is assessed via Abs-MPJPE, MPJPE, PA, EPE$_{\mathrm{2D}}$, and Jitter, quantifying absolute, wrist-relative, and Procrustes-aligned 3D joint errors, 2D projection error, and temporal smoothness, respectively. Finally, world-space reconstruction is measured with W-MPJPE, WA-MPJPE, RTE, and Accel, corresponding to world-space and aligned 3D joint errors, normalized wrist trajectory error, and acceleration error, respectively.

\paragraph{Baselines.}
For camera-space reconstruction, we compare with InterWild~\citep{moon2023interwild}, HaMeR~\citep{pavlakos2024hamer}, Hamba~\citep{dong2024hamba}, WildHands~\citep{prakash2024wildhands}, OmniHands~\citep{lin2024omnihands}, WiLoR~\citep{potamias2024wilor}, Dyn-HaMR~\citep{yu2025dynhamr}, HaWoR~\citep{zhang2025hawor}, ViDiHand~\citep{wang2026vidihand}, MINT~\citep{zhu2026mint}, and ACE-Ego-Hand~\citep{liu2026ace}. For world-space reconstruction, we compare with HaWoR, Dyn-HaMR, MINT, WiLoR-SLAM, and ACE-SLAM. WiLoR-SLAM and ACE-SLAM transform WiLoR and ACE-Ego-Hand predictions, respectively, into world coordinates using shared DROID-SLAM~\citep{teed2021droidslam} camera poses scaled by Metric3D v2~\citep{hu2024metric3dv2}.

\paragraph{Implementation details.}
Both training stages use the AdamW~\citep{loshchilov2019adamw} optimizer with a weight decay of $0.01$, a global batch size of 64, linear warmup with cosine annealing, and global gradient norm clipping at 1.0. Specifically, Stage I trains for 300k steps on 4-frame clips with a learning rate of $5\times10^{-5}$ for the mask and MANO heads. Stage II then trains for 100k steps under the 36-frame $4+16\times 2$ temporal layout, setting the learning rate to $3\times10^{-5}$ for the camera head and $1\times10^{-5}$ for all other trainable modules. Across both stages, full-resolution inputs are processed at $378\times 518$, while hand bounding boxes are expanded by $1.5\times$ and resized to $256\times 256$ for the crop branch.

\begin{figure*}[t]
\centering
\includegraphics[width=\textwidth]
{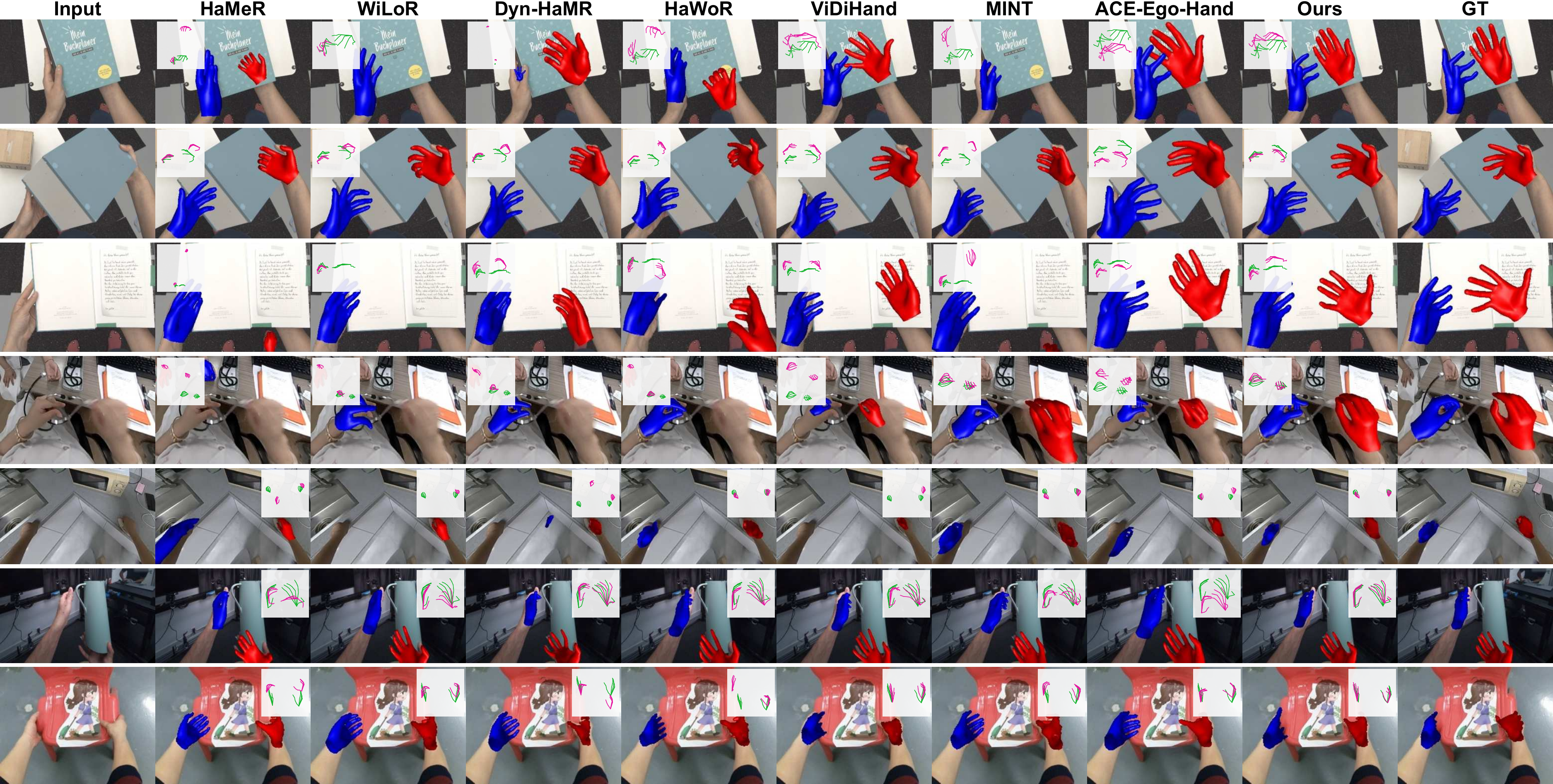}
\caption{\textbf{Camera-space qualitative comparison.} InfiniHand recovers both hands without missed detections while producing precise hand motion under severe occlusion and in-the-wild scenes. Insets in the upper corners compare ground-truth (green) and predicted (magenta) joints in the camera $X$--$Z$ plane.}
\label{fig:camera_qualitative}
\end{figure*}

\begin{table*}[t]
\centering
\caption{\textbf{World-space quantitative comparison.} We evaluate world-space hand motion and trajectory metrics across three benchmarks, with InfiniHand achieving the lowest W-MPJPE.}
\label{tab:world_main}
\begingroup
\setlength{\tabcolsep}{2pt}
\renewcommand{\arraystretch}{1.08}
\scriptsize
\resizebox{\textwidth}{!}{%
\begin{tabular}{@{}l*{4}{c}@{\hspace{7pt}}*{4}{c}@{\hspace{7pt}}*{4}{c}@{}}
\toprule
Method & \multicolumn{4}{c}{\textbf{ARCTIC}} & \multicolumn{4}{c}{\textbf{HOT3D}} & \multicolumn{4}{c}{\textbf{Gen-HumanEgo}} \\
\cmidrule(lr){2-5}\cmidrule(lr){6-9}\cmidrule(lr){10-13}

 & W-MPJPE$\downarrow$ & WA-MPJPE$\downarrow$ & RTE$\downarrow$ & Accel$\downarrow$ & W-MPJPE$\downarrow$ & WA-MPJPE$\downarrow$ & RTE$\downarrow$ & Accel$\downarrow$ & W-MPJPE$\downarrow$ & WA-MPJPE$\downarrow$ & RTE$\downarrow$ & Accel$\downarrow$ \\
\midrule
WiLoR-SLAM & 65.86 & 46.31 & 5.488 & 21.38 & 106.77 & 45.20 & 9.764 & 16.08 & 97.53 & 33.32 & 3.79 & 5.01 \\
ACE-SLAM & 63.27 & 36.36 & 4.076 & \underline{7.16} & 89.87 & 38.47 & 6.383 & 9.64 & 64.62 & 29.67 & 2.23 & 2.94 \\
HaWoR & 95.39 & 45.45 & 3.219 & 15.02 & 93.70 & 35.02 & 6.175 & 12.01 & 127.93 & 48.09 & 7.11 & 8.78 \\
Dyn-HaMR & 114.10 & 63.60 & 3.790 & 13.60 & 296.72 & 132.08 & 10.777 & 13.90 & 175.91 & 80.43 & 9.48 & 9.17 \\
MINT & 108.42 & 45.01 & 4.959 & 12.48 & 99.34 & 38.10 & 5.338 & 10.20 & 65.29 & 28.63 & 2.75 & 2.35 \\
\midrule
Ours & \underline{46.79} & \underline{22.77} & \underline{1.351} & \textbf{5.34} & \underline{81.05} & \underline{29.06} & \underline{4.64} & \underline{7.40} & \underline{48.68} & \underline{25.59} & \underline{1.92} & \textbf{2.09} \\
Ours + BA & \textbf{43.83} & \textbf{22.14} & \textbf{1.304} & \textbf{5.34} & \textbf{69.60} & \textbf{25.51} & \textbf{4.02} & \textbf{6.86} & \textbf{46.36} & \textbf{25.17} & \textbf{1.90} & \underline{2.10} \\
\bottomrule
\end{tabular}}
\endgroup
\end{table*}

\begin{figure*}[t]
\centering
\includegraphics[width=\textwidth]{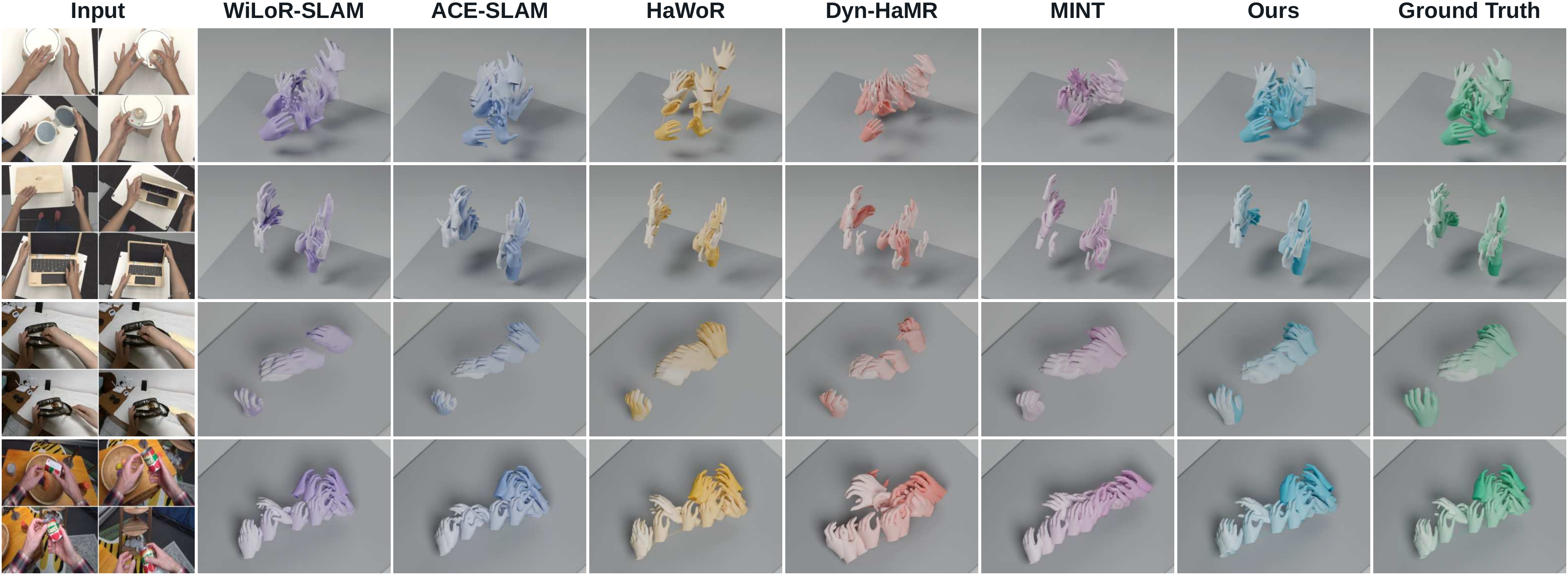}
\caption{\textbf{World-space qualitative comparison.} Visual results on diverse datasets show that our reconstructed articulation and trajectories achieve the highest fidelity to ground truth.}
\label{fig:world_qualitative}
\end{figure*}

\subsection{Camera-space results analysis}
\label{sec:comparison}
Table~\ref{tab:camera_main} shows that InfiniHand improves both absolute hand placement and articulated geometry across in-domain and out-of-domain benchmarks. On ARCTIC, it achieves a PA of 6.14~mm compared with 7.71~mm for ACE-Ego-Hand~\citep{liu2026ace}, reflecting more accurate hand geometry. The gains extend to absolute placement on unseen data, with 25.2\% lower Abs-MPJPE than WiLoR~\citep{potamias2024wilor} on HOI4D. Higher FAcc and lower Jitter further support reliable hand detection and temporally stable predictions. As shown in Fig.~\ref{fig:camera_qualitative}, InfiniHand better preserves finger articulation during hand--object interactions and aligns the reconstructed meshes with the visible hands under occlusion and clutter. These qualitative improvements agree with the lower 2D projection error, demonstrating that the reconstructed geometry remains consistent with the image evidence.

\subsection{World-space results analysis}
\label{sec:world_comparison}
Table~\ref{tab:world_main} demonstrates that InfiniHand improves both global hand placement and motion reconstruction. With sparse BA, it achieves a W-MPJPE of 43.83~mm compared with 108.42~mm for MINT~\citep{zhu2026mint} on ARCTIC, and reduces W-MPJPE by 63.8\% relative to HaWoR~\citep{zhang2025hawor} on Gen-HumanEgo. The lower WA-MPJPE shows that the advantage persists after sequence-level alignment, while the lower RTE indicates more accurate trajectories. Even without BA, the streaming predictions outperform all baselines in these metrics; sparse BA further improves global consistency by revisiting cross-frame geometric constraints. Low acceleration error further indicates that the method preserves motion dynamics alongside spatial accuracy. Fig.~\ref{fig:world_qualitative} highlights these improvements in world-space reconstruction, demonstrating that InfiniHand consistently preserves realistic hand scale, relative inter-hand spacing, and natural curved trajectories, whereas competing baselines exhibit noticeable hand drift or distorted motion paths.

\subsection{Efficiency analysis}
\label{sec:efficiency}
InfiniHand achieves a prediction throughput of 11.90~FPS, more than twice that of HaWoR~\citep{zhang2025hawor} (5.48~FPS) and faster than WiLoR-SLAM~\citep{potamias2024wilor,teed2021droidslam} (8.53~FPS), ACE-SLAM (3.11~FPS), and Dyn-HaMR~\citep{yu2025dynhamr} (0.81~FPS). Although MINT~\citep{zhu2026mint} (21.44~FPS) runs faster, its W-MPJPE is substantially higher than ours on all three benchmarks (Table~\ref{tab:world_main}), showing our favorable balance between reconstruction accuracy and efficiency. This efficiency comes from reusing streaming features and overlapping feature extraction with prediction decoding. One worker processes the upcoming 16-frame window using four-anchor context, while another decodes hand and camera predictions for the preceding window, reducing idle time between the two stages. A third worker performs sparse BA in the background whenever the binary keyframe pool exceeds its capacity threshold.

\subsection{Ablation Study}
\label{sec:egodex_ablation}
Table~\ref{tab:egodex_ablation} evaluates the main components on all 34 ARCTIC~\citep{fan2023arctic} test scenes. (1) Hand-centered appearance features complement the geometric representation: removing WiLoR features increases PA from 6.14 to 9.01~mm and degrades image alignment. (2) LSP primarily improves absolute hand placement; its removal increases absolute joint error while causing smaller changes in local articulation. (3) Stage II strengthens temporal stability as well as reconstruction accuracy, with Jitter rising from 2.35 to 5.38 when it is omitted. (4) Removing GCA memory likewise worsens spatial accuracy and temporal stability, supporting the benefit of persistent geometric context across windows. (5) Replacing mask-guided localization with DINO bounding boxes reduces hand detection accuracy and increases EPE$_{\mathrm{2D}}$ from 7.65 to 15.42 pixels. Despite slightly lower Jitter, this variant produces less accurate hand geometry and image alignment, underscoring the importance of reliable localization for reconstruction.

\section{Limitations}
\label{sec:limitations}

\begin{table}[t]
\centering
\caption{\textbf{Quantitative ablation study on ARCTIC.} We systematically evaluate the impact of our feature fusion modules, spatiotemporal memory, and two-stage training strategy.}
\label{tab:egodex_ablation}
\begingroup
\setlength{\tabcolsep}{1pt}
\renewcommand{\arraystretch}{1.08}
\fontsize{6.5}{7.4}\selectfont
\resizebox{\columnwidth}{!}{%
\begin{tabular}{@{}l*{6}{c}@{}}
\toprule
Variant & FAcc$\uparrow$ & Abs-MPJPE$\downarrow$ & MPJPE$\downarrow$ & PA$\downarrow$ & EPE$_{\mathrm{2D}}\downarrow$ & Jitter$\downarrow$ \\
\midrule
w/o WiLoR Features & 0.997 & 47.39 & 19.72 & 9.01 & 13.95 & 3.27 \\
w/o LSP & 0.997 & 35.93 & 15.48 & 6.28 & 7.97 & 2.56 \\
w/o Stage II & 0.974 & 40.90 & 17.06 & 7.22 & 11.81 & 5.38 \\
w/o GCA Memory & 0.995 & 37.66 & 16.35 & 6.53 & 8.82 & 4.31 \\
w/ DINO bbox & 0.906 & 45.73 & 19.32 & 8.75 & 15.42 & 2.18 \\
\midrule
\textbf{Full Model} & 0.997 & 32.96 & 15.12 & 6.14 & 7.65 & 2.35 \\
\bottomrule
\end{tabular}}
\endgroup
\end{table}

While InfiniHand achieves accurate world-space 3D hand reconstruction, there remains clear room for further improvement across several key technical aspects. In terms of scale recovery, the underlying LingBot-Map framework lacks inherent metric scale, requiring an auxiliary post-processing alignment model whose downstream estimation errors can inevitably propagate to predicted hand positions and global motion trajectories. Regarding data coverage, high-quality egocentric datasets featuring complex, large-amplitude two-hand interactions and camera dynamics remain scarce, restricting generalization to unconstrained in-the-wild sequences with rapid viewpoint changes, severe occlusions, and intermittent hand visibility. On the supervision front, monocular reconstruction pipelines often inherit time-varying scale drift from derived monocular annotations, creating temporally inconsistent training targets that compromise long-sequence trajectory accuracy and temporal smoothness despite global scale alignment.
\section{Conclusion}
\label{sec:conclusion}
We present InfiniHand, a streaming feed-forward framework that reconstructs articulated hands while predicting hand locations and camera trajectories from egocentric video. A two-stage training scheme first learns camera-space hand reconstruction and then performs joint hand and camera motion estimation over longer sequences, with sparse keyframe bundle adjustment improving global trajectory consistency. Extensive evaluations across in-domain benchmarks and out-of-domain videos demonstrate strong reconstruction quality and generalization. Beyond accuracy, the streaming design achieves more than twice the inference throughput of HaWoR under standard timing protocols. Collectively, these advances enable faster and more reliable 3D annotation of egocentric video, unlocking downstream applications in dexterous data augmentation, human-to-robot motion transfer, and manipulation policy learning. By converting abundant human video into structured 3D hand-motion supervision, InfiniHand offers a scalable path toward data generation for embodied intelligence.
\section*{LLM Usage Statement}
\label{sec:statement}

In this work, we used generative AI tools solely to assist with translation, literature summarization and analysis, and manuscript editing to enhance readability. Conversely, we did not use generative AI tools for designing research methods and experiments, implementing methodologies, interpreting results, proposing or refining hypotheses, cleaning and reformatting datasets, or supporting qualitative and thematic data analysis. Generating synthetic datasets, proposing mathematical claims, providing key elements for proofs, and assisting in writing proofs are not applicable to this work. We thoroughly reviewed all AI-assisted outputs by manually cross-checking translated and polished text sentence by sentence to ensure the original intent remained uncompromised. Finally, we take full responsibility for the final content of this work, including all text, claims, and artifacts produced with the aid of generative AI.

{
    \small
    \bibliographystyle{ieeenat_fullname}
    \bibliography{main,supplement}

@article{potamias2024wilor,
  title={{WiLoR}: End-to-End 3D Hand Localization and Reconstruction in the Wild},
  author={Potamias, Rolandos Alexandros and Zhang, Jinglei and Deng, Jiankang and Zafeiriou, Stefanos},
  journal={arXiv preprint arXiv:2409.12259},
  year={2024}
}

@article{zhang2025hawor,
  title={{HaWoR}: World-Space Hand Motion Reconstruction from Egocentric Videos},
  author={Zhang, Jinglei and Deng, Jiankang and Ma, Chao and Potamias, Rolandos Alexandros},
  journal={arXiv preprint arXiv:2501.02973},
  year={2025}
}

@article{wang2026vidihand,
  title={The Surprising Effectiveness of Video Diffusion Models for Hand Motion Reconstruction},
  author={Wang, Yuxi and Jin, Chengkai and Liu, Yufei and Ouyang, Wenqi and Wei, Tianyi and Zeng, Zhiwei and Huang, Siyuan and Shen, Zhiqi and Pan, Xingang},
  journal={arXiv preprint arXiv:2606.30308},
  year={2026}
}

@article{zhang2026loger,
  title={{LoGeR}: Long-Context Geometric Reconstruction with Hybrid Memory},
  author={Zhang, Junyi and Herrmann, Charles and Hur, Junhwa and Sun, Chen and Yang, Ming-Hsuan and Cole, Forrester and Darrell, Trevor and Sun, Deqing},
  journal={arXiv preprint arXiv:2603.03269},
  year={2026}
}

@article{chen2026lingbotmap,
  title={{LingBot-Map}: Geometric Context Transformer for Streaming 3D Reconstruction},
  author={Chen, Lin-Zhuo and Gao, Jian and Zhang, Shangzhan and Chen, Yihang and Cheng, Ka Leong and Sun, Yipengjing and Hu, Liangxiao and Xue, Nan and Zhu, Xing and Shen, Yujun and Yao, Yao and Xu, Yinghao},
  journal={arXiv preprint arXiv:2604.14141},
  year={2026}
}

@article{teed2021droidslam,
  title={{DROID-SLAM}: Deep Visual {SLAM} for Monocular, Stereo, and {RGB-D} Cameras},
  author={Teed, Zachary and Deng, Jia},
  journal={Advances in Neural Information Processing Systems},
  volume={34},
  year={2021}
}

@article{murartal2017orbslam2,
  title={{ORB-SLAM2}: An Open-Source {SLAM} System for Monocular, Stereo, and {RGB-D} Cameras},
  author={Mur-Artal, Ra{\'u}l and Tard{\'o}s, Juan D.},
  journal={IEEE Transactions on Robotics},
  volume={33},
  number={5},
  pages={1255--1262},
  year={2017}
}

@inproceedings{fan2023arctic,
  title={{ARCTIC}: A Dataset for Dexterous Bimanual Hand-Object Manipulation},
  author={Fan, Zicong and Taheri, Omid and Tzionas, Dimitrios and Kocabas, Muhammed and Kaufmann, Manuel and Black, Michael J. and Hilliges, Otmar},
  booktitle={Proceedings of the IEEE/CVF Conference on Computer Vision and Pattern Recognition},
  year={2023}
}

@inproceedings{banerjee2025hot3d,
  title={{HOT3D}: Hand and Object Tracking in 3D from Egocentric Multi-View Videos},
  author={Banerjee, Prithviraj and Shkodrani, Sindi and Moulon, Pierre and Hampali, Shreyas and Han, Shangchen and Zhang, Fan and Zhang, Linguang and Fountain, Jade and Miller, Edward and Basol, Selen and Newcombe, Richard and Wang, Robert and Engel, Jakob Julian and Hodan, Tomas},
  booktitle={Proceedings of the IEEE/CVF Conference on Computer Vision and Pattern Recognition},
  year={2025}
}

@article{hoque2025egodex,
  title={{EgoDex}: Learning Dexterous Manipulation from Large-Scale Egocentric Video},
  author={Hoque, Ryan and Huang, Peide and Yoon, David J. and Sivapurapu, Mouli and Zhang, Jian},
  journal={arXiv preprint arXiv:2505.11709},
  year={2025}
}

@inproceedings{hampali2020honnotate,
  title={{HOnnotate}: A Method for 3D Annotation of Hand and Object Poses},
  author={Hampali, Shreyas and Rad, Mahdi and Oberweger, Markus and Lepetit, Vincent},
  booktitle={Proceedings of the IEEE/CVF Conference on Computer Vision and Pattern Recognition},
  year={2020}
}

@article{romero2017mano,
  title={Embodied Hands: Modeling and Capturing Hands and Bodies Together},
  author={Romero, Javier and Tzionas, Dimitrios and Black, Michael J.},
  journal={ACM Transactions on Graphics},
  volume={36},
  number={6},
  pages={245:1--245:17},
  year={2017}
}

@article{zhu2026mint,
  title={Mint: A unified model for world-space camera and hand motion estimation from scalable egocentric pipeline supervision},
  author={Zhu, Zijie and Cai, Weiren and Wang, Yizhou and Yang, Zhenjie and Liu, Yide and Chen, Jiahao and He, Guanqi},
  journal={arXiv preprint arXiv:2609.04958},
  year={2026}
}

@inproceedings{pavlakos2024hamer,
 title={Reconstructing Hands in 3D with Transformers},
 author={Pavlakos, Georgios and Shan, Dandan and Radosavovic, Ilija and Kanazawa, Angjoo and Fouhey, David and Malik, Jitendra},
 booktitle={Proceedings of the IEEE/CVF Conference on Computer Vision and Pattern Recognition},
 year={2024}
}

@inproceedings{yu2025dynhamr,
 title={{Dyn-HaMR}: Recovering 4D Interacting Hand Motion from a Dynamic Camera},
 author={Yu, Zhengdi and Zafeiriou, Stefanos and Birdal, Tolga},
 booktitle={Proceedings of the IEEE/CVF Conference on Computer Vision and Pattern Recognition},
 year={2025}
}

@article{li2026egowam,
  title={{EgoWAM}: World Action Models Beyond Pixels with In-the-Wild Egocentric Human Data},
  author={Li, Baoyu and Yin, Xinchen and Lin, Mengying and Zhang, Yixin and Xu, Danfei},
  journal={arXiv preprint arXiv:2607.08436},
  year={2026}
}

@inproceedings{papagiannis2025rx,
  title={{R+X}: Retrieval and Execution from Everyday Human Videos},
  author={Papagiannis, Georgios and Di Palo, Norman and Vitiello, Pietro and Johns, Edward},
  booktitle={IEEE International Conference on Robotics and Automation},
  year={2025}
}

@article{wang2025vggt,
  author       = {Jianyuan Wang and
                  Minghao Chen and
                  Nikita Karaev and
                  Andrea Vedaldi and
                  Christian Rupprecht and
                  David Novotn{\'{y}}},
  title        = {{VGGT:} Visual Geometry Grounded Transformer},
  journal      = {CoRR},
  volume       = {abs/2503.11651},
  year         = {2025},
  url          = {https://doi.org/10.48550/arXiv.2503.11651},
  doi          = {10.48550/ARXIV.2503.11651},
  eprinttype   = {arXiv},
  eprint       = {2503.11651},
  bibsource    = {dblp computer science bibliography, https://dblp.org}
}

@inproceedings{lin2021meshgraphormer,
  author       = {Kevin Lin and
                  Lijuan Wang and
                  Zicheng Liu},
  title        = {Mesh Graphormer},
  booktitle    = {2021 {IEEE/CVF} International Conference on Computer Vision, {ICCV}
                  2021, Montreal, QC, Canada, October 10-17, 2021},
  pages        = {12919--12928},
  publisher    = {{IEEE}},
  year         = {2021},
  url          = {https://doi.org/10.1109/ICCV48922.2021.01270},
  doi          = {10.1109/ICCV48922.2021.01270},
  bibsource    = {dblp computer science bibliography, https://dblp.org}
}

@inproceedings{lin2021metro,
  author       = {Kevin Lin and
                  Lijuan Wang and
                  Zicheng Liu},
  title        = {End-to-End Human Pose and Mesh Reconstruction with Transformers},
  booktitle    = {{IEEE} Conference on Computer Vision and Pattern Recognition, {CVPR}
                  2021, virtual, June 19-25, 2021},
  pages        = {1954--1963},
  publisher    = {Computer Vision Foundation / {IEEE}},
  year         = {2021},
  url          = {https://openaccess.thecvf.com/content/CVPR2021/html/Lin\_End-to-End\_Human\_Pose\_and\_Mesh\_Reconstruction\_with\_Transformers\_CVPR\_2021\_paper.html},
  doi          = {10.1109/CVPR46437.2021.00199},
  bibsource    = {dblp computer science bibliography, https://dblp.org}
}

@inproceedings{leroy2024mast3r,
  author       = {Vincent Leroy and
                  Yohann Cabon and
                  J{\'{e}}r{\^{o}}me Revaud},
  editor       = {Ales Leonardis and
                  Elisa Ricci and
                  Stefan Roth and
                  Olga Russakovsky and
                  Torsten Sattler and
                  G{\"{u}}l Varol},
  title        = {Grounding Image Matching in 3D with MASt3R},
  booktitle    = {Computer Vision - {ECCV} 2024 - 18th European Conference, Milan, Italy,
                  September 29-October 4, 2024, Proceedings, Part {LXXII}},
  series       = {Lecture Notes in Computer Science},
  volume       = {15130},
  pages        = {71--91},
  publisher    = {Springer},
  year         = {2024},
  url          = {https://doi.org/10.1007/978-3-031-73220-1\_5},
  doi          = {10.1007/978-3-031-73220-1\_5},
  bibsource    = {dblp computer science bibliography, https://dblp.org}
}

@article{wang2025cut3r,
  author       = {Qianqian Wang and
                  Yifei Zhang and
                  Aleksander Holynski and
                  Alexei A. Efros and
                  Angjoo Kanazawa},
  title        = {Continuous 3D Perception Model with Persistent State},
  journal      = {CoRR},
  volume       = {abs/2501.12387},
  year         = {2025},
  url          = {https://doi.org/10.48550/arXiv.2501.12387},
  doi          = {10.48550/ARXIV.2501.12387},
  eprinttype   = {arXiv},
  eprint       = {2501.12387},
  bibsource    = {dblp computer science bibliography, https://dblp.org}
}

@inproceedings{wang2025spann3r,
  author       = {Hengyi Wang and
                  Lourdes Agapito},
  title        = {3D Reconstruction with Spatial Memory},
  booktitle    = {International Conference on 3D Vision, 3DV 2025, Singapore, March
                  25-28, 2025},
  pages        = {78--89},
  publisher    = {{IEEE}},
  year         = {2025},
  url          = {https://doi.org/10.1109/3DV66043.2025.00013},
  doi          = {10.1109/3DV66043.2025.00013},
  bibsource    = {dblp computer science bibliography, https://dblp.org}
}

@article{dong2024hamba,
  title={{Hamba: Single-view 3D Hand Reconstruction with Graph-guided Bi-Scanning Mamba}},
  author={Dong, Haoye and Chharia, Aviral and Gou, Wenbo and Carrasco, Francisco Vicente and De la Torre, Fernando},
  journal={arXiv preprint arXiv:2407.09646},
  year={2024}
}

@article{prakash2024wildhands,
  title={{3D Hand Pose Estimation in Everyday Egocentric Images}},
  author={Prakash, Aditya and Tu, Ruisen and Chang, Matthew and Gupta, Saurabh},
  journal={arXiv preprint arXiv:2312.06583},
  year={2023}
}

@article{lin2024omnihands,
  title={{OmniHands: Towards Robust 4D Hand Mesh Recovery via A Versatile Transformer}},
  author={Lin, Dixuan and Zhang, Yuxiang and Li, Mengcheng and Jing, Wei and Yan, Qi and Wang, Qianying and Liu, Yebin and Zhang, Hongwen},
  journal={arXiv preprint arXiv:2405.20330},
  year={2024}
}

@article{moon2023interwild,
  title={{Bringing Inputs to Shared Domains for 3D Interacting Hands Recovery in the Wild}},
  author={Moon, Gyeongsik},
  journal={arXiv preprint arXiv:2303.13652},
  year={2023}
}

@article{murai2025mast3rslam,
  title={{MASt3R-SLAM: Real-Time Dense SLAM with 3D Reconstruction Priors}},
  author={Murai, Riku and Dexheimer, Eric and Davison, Andrew J.},
  journal={arXiv preprint arXiv:2412.12392},
  year={2024}
}

@article{maggio2025vggtslam,
  title={{VGGT-SLAM: Dense RGB SLAM Optimized on the SL(4) Manifold}},
  author={Maggio, Dominic and Lim, Hyungtae and Carlone, Luca},
  journal={arXiv preprint arXiv:2505.12549},
  year={2025}
}

@article{ren2026m3,
  title={{M$^3$: Dense Matching Meets Multi-View Foundation Models for Monocular Gaussian Splatting SLAM}},
  author={Ren, Kerui and Li, Guanghao and Jiang, Changjian and Xu, Yingxiang and Lu, Tao and Xu, Linning and Dong, Junting and Pang, Jiangmiao and Yu, Mulin and Dai, Bo},
  journal={arXiv preprint arXiv:2603.16844},
  year={2026}
}

@article{yin2023metric3d,
  title={{Metric3D: Towards Zero-shot Metric 3D Prediction from A Single Image}},
  author={Yin, Wei and Zhang, Chi and Chen, Hao and Cai, Zhipeng and Yu, Gang and Wang, Kaixuan and Chen, Xiaozhi and Shen, Chunhua},
  journal={arXiv preprint arXiv:2307.10984},
  year={2023}
}

@inproceedings{liu2022hoi4d,
 title={{HOI4D: A 4D Egocentric Dataset for Category-Level Human-Object Interaction}},
 author={Liu, Yunze and Liu, Yun and Jiang, Che and Lyu, Kangbo and Wan, Weikang and Shen, Hao and Liang, Boqiang and Fu, Zhoujie and Wang, He and Yi, Li},
 booktitle={Proceedings of the IEEE/CVF Conference on Computer Vision and Pattern Recognition},
 year={2022}
}

@inproceedings{grauman2022ego4d,
 title={{Ego4D: Around the World in 3,000 Hours of Egocentric Video}},
 author={Grauman, Kristen and Westbury, Andrew and Byrne, Eugene and Chavis, Zachary and Furnari, Antonino and Girdhar, Rohit and Hamburger, Jackson and Jiang, Hao and Liu, Miao and Liu, Xingyu and Martin, Miguel and Nagarajan, Tushar and Radosavovic, Ilija and Ramakrishnan, Santhosh Kumar and Ryan, Fiona and Sharma, Jayant and Wray, Michael and Xu, Mengmeng and Xu, Eric Zhongcong and Zhao, Chen and Bansal, Siddhant and Batra, Dhruv and Cartillier, Vincent and Crane, Sean and Do, Tien and Doulaty, Morrie and Erapalli, Akshay and Feichtenhofer, Christoph and Fragomeni, Adriano and Fu, Qichen and Gebreselasie, Abrham and Gonzalez, Cristina and Hillis, James and Huang, Xuhua and Huang, Yifei and Jia, Wenqi and Khoo, Weslie and Kolar, Jachym and Kottur, Satwik and Kumar, Anurag and Landini, Federico and Li, Chao and Li, Yanghao and Li, Zhenqiang and Mangalam, Karttikeya and Modhugu, Raghava and Munro, Jonathan and Murrell, Tullie and Nishiyasu, Takumi and Price, Will and Puentes, Paola Ruiz and Ramazanova, Merey and Sari, Leda and Somasundaram, Kiran and Southerland, Audrey and Sugano, Yusuke and Tao, Ruijie and Vo, Minh and Wang, Yuchen and Wu, Xindi and Yagi, Takuma and Zhao, Ziwei and Zhu, Yunyi and Arbelaez, Pablo and Crandall, David and Damen, Dima and Farinella, Giovanni Maria and Fuegen, Christian and Ghanem, Bernard and Ithapu, Vamsi Krishna and Jawahar, C. V. and Joo, Hanbyul and Kitani, Kris and Li, Haizhou and Newcombe, Richard and Oliva, Aude and Park, Hyun Soo and Rehg, James M. and Sato, Yoichi and Shi, Jianbo and Shou, Mike Zheng and Torralba, Antonio and Torresani, Lorenzo and Yan, Mingfei and Malik, Jitendra},
 booktitle={{IEEE/CVF} Conference on Computer Vision and Pattern Recognition,
            {CVPR} 2022, New Orleans, LA, USA, June 18-24, 2022},
 pages={18973--18990},
 publisher={{IEEE}},
 year={2022},
 url={https://doi.org/10.1109/CVPR52688.2022.01842},
 doi={10.1109/CVPR52688.2022.01842},
 bibsource={dblp computer science bibliography, https://dblp.org}
}

@article{hu2024metric3dv2,
 title={{Metric3Dv2: A Versatile Monocular Geometric Foundation Model for Zero-shot Metric Depth and Surface Normal Estimation}},
 author={Hu, Mu and Yin, Wei and Zhang, Chi and Cai, Zhipeng and Long, Xiaoxiao and Wang, Kaixuan and Chen, Hao and Yu, Gang and Shen, Chunhua and Shen, Shaojie},
 journal={arXiv preprint arXiv:2404.15506},
 year={2024}
}

@inproceedings{chao2021dexycb,
 title={{DexYCB}: A Benchmark for Capturing Hand Grasping of Objects},
 author={Chao, Yu-Wei and Yang, Wei and Xiang, Yu and Molchanov, Pavlo and Handa, Ankur and Tremblay, Jonathan and Narang, Yashraj S. and {Van Wyk}, Karl and Iqbal, Umar and Birchfield, Stan and Kautz, Jan and Fox, Dieter},
 booktitle={Proceedings of the IEEE/CVF Conference on Computer Vision and Pattern Recognition},
 year={2021}
}

@inproceedings{hampali2022keypointtransformer,
 title={Keypoint Transformer: Solving Joint Identification in Challenging Hands and Object Interactions for Accurate 3D Pose Estimation},
 author={Hampali, Shreyas and Deb Sarkar, Sayan and Rad, Mahdi and Lepetit, Vincent},
 booktitle={Proceedings of the IEEE/CVF Conference on Computer Vision and Pattern Recognition},
 year={2022}
}

@article{punamiya2026egoverse,
 title={{EgoVerse}: An Egocentric Human Dataset for Robot Learning from Around the World},
 author={Punamiya, Ryan and Kareer, Simar and Liu, Zeyi and Citron, Josh and Qiu, Ri-Zhao and Cai, Xiongyi and Gavryushin, Alexey and Chen, Jiaqi and Liconti, Davide and Zhu, Lawrence Y. and Aphiwetsa, Patcharapong and Li, Baoyu and Cheluva, Aniketh and Kuppili, Pranav and Liu, Yangcen and Patel, Dhruv and Gao, Aidan and Chung, Hye-Young and Co, Ryan and Zbizika, Renee and Liu, Jeff and Xu, Xiaomeng and Xiong, Haoyu and Chen, Geng and Oliani, Sebastiano and Xuan, Wenkai and Yang, Chenyu and Wang, Xi and Fort, James and Newcombe, Richard and Gao, Josh and Chong, Jason and Matsuda, Garrett and Doriwala, Aseem and Pollefeys, Marc and Katzschmann, Robert and Wang, Xiaolong and Song, Shuran and Hoffman, Judy and Xu, Danfei},
 journal={arXiv preprint arXiv:2604.07607},
 year={2026}
}

@inproceedings{ranftl2021dpt,
 title={Vision Transformers for Dense Prediction},
 author={Ranftl, Ren{\'e} and Bochkovskiy, Alexey and Koltun, Vladlen},
 booktitle={Proceedings of the IEEE/CVF International Conference on Computer Vision},
 year={2021}
}

@inproceedings{lin2017refinenet,
 title={{RefineNet}: Multi-Path Refinement Networks for High-Resolution Semantic Segmentation},
 author={Lin, Guosheng and Milan, Anton and Shen, Chunhua and Reid, Ian},
 booktitle={Proceedings of the IEEE Conference on Computer Vision and Pattern Recognition},
 year={2017}
}

@inproceedings{loshchilov2019adamw,
 title={Decoupled Weight Decay Regularization},
 author={Loshchilov, Ilya and Hutter, Frank},
 booktitle={International Conference on Learning Representations},
 year={2019}
}

@article{liu2026ace,
  title={ACE-Ego-Hand: Repurposing Video Diffusion Models for Occlusion-Robust Egocentric 3D Hand Motion Recovery},
  author={Liu, Yufei and Wang, Xixi and Li, Hao and Zhao, Ganlong and Cai, Kaitong and Jin, Chengkai and Liu, Chunxiao and Liu, Jianbo and Huang, Siyuan and Pan, Xingang and others},
  journal={arXiv preprint arXiv:2608.20308},
  year={2026}
}

@misc{egostandard,
  title        = {{EgoStandard}: Hardware-Assisted Egocentric Multi-Modal Hand Motion Benchmark},
  author       = {{Lightwheel AI}},
  howpublished = {\url{https://lightwheel.ai/egosuite}},
  year         = {2025},
}

@misc{genrobot2026genhumanego,
  title        = {{Gen-HumanEgo}},
  author       = {{GenRobot}},
  howpublished = {\url{https://huggingface.co/datasets/genrobot2025/Gen-HumanEgo}},
  year         = {2026},
}

@misc{egohandbench_supp,
    author = {{EgoHandBench}},
    title = {{EgoHandBench}: A Unified Benchmark for Egocentric {3D} Hand Reconstruction},
    howpublished = {Benchmark project and leaderboard},
    year = {2026},
}

@inproceedings{kwon2021h2o,
    author = {Taein Kwon and Bugra Tekin and Jan St{\"u}hmer and Federica Bogo and Marc Pollefeys},
    title = {{H2O}: Two Hands Manipulating Objects for First Person Interaction Recognition},
    booktitle = {ICCV},
    pages = {10118--10128},
    year = {2021},
}

@inproceedings{zhan2024oakink2,
    author = {Xinyu Zhan and Lixin Yang and Yifei Zhao and Kangrui Mao and Hanlin Xu and Zenan Lin and Kailin Li and Cewu Lu},
    title = {{OakInk2}: A Dataset of Bimanual Hands-Object Manipulation in Complex Task Completion},
    booktitle = {CVPR},
    pages = {445--456},
    year = {2024},
}

@article{carion2025sam3,
    author = {Nicolas Carion and Laura Gustafson and Yuan-Ting Hu and Shoubhik Debnath and Ronghang Hu and Didac Suris and Chaitanya Ryali and Kalyan Vasudev Alwala and Haitham Khedr and Andrew Huang and Jie Lei and Tengyu Ma and Baishan Guo and Arpit Kalla and Markus Marks and Joseph Greer and Meng Wang and Peize Sun and Roman R{\"a}dle and Triantafyllos Afouras and Effrosyni Mavroudi and Katherine Xu and Tsung-Han Wu and Yu Zhou and Liliane Momeni and Rishi Hazra and Shuangrui Ding and Sagar Vaze and Francois Porcher and Feng Li and Siyuan Li and Aishwarya Kamath and Ho Kei Cheng and Piotr Doll{\'a}r and Nikhila Ravi and Kate Saenko and Pengchuan Zhang and Christoph Feichtenhofer},
    title = {{SAM 3}: Segment Anything with Concepts},
    journal = {arXiv preprint arXiv:2511.16719},
    year = {2025},
}

@inproceedings{xie2026egohandicl,
    author = {Binzhu Xie and Shi Qiu and Sicheng Zhang and Yinqiao Wang and Hao Xu and Muzammal Naseer and Chi-Wing Fu and Pheng-Ann Heng},
    title = {{EgoHandICL}: Egocentric {3D} Hand Reconstruction with In-Context Learning},
    booktitle = {ICLR},
    year = {2026},
}

@article{xu2026handflow,
    author = {Mingxi Xu and Bowen Duan and Yi Gu and Zhengyang Shen and Renjing Xu and Yutao Yue},
    title = {{HandFlow}: Fully Generative {4D} Hand Recovery with Flow Matching},
    journal = {arXiv preprint arXiv:2607.11221},
    year = {2026},
}

@article{si2026anyhand,
    author = {Chen Si and Yulin Liu and Bo Ai and Jianwen Xie and Rolandos Alexandros Potamias and Chuanxia Zheng and Hao Su},
    title = {{AnyHand}: A Large-Scale Synthetic Dataset for {RGB(-D)} Hand Pose Estimation},
    journal = {arXiv preprint arXiv:2603.25726},
    year = {2026},
}

@article{ismayilzada2026padhand,
    author = {Elkhan Ismayilzada and Yufei Zhang and Zijun Cui},
    title = {{PAD-Hand}: Physics-Aware Diffusion for Hand Motion Recovery},
    journal = {arXiv preprint arXiv:2603.26068},
    year = {2026},
}

@inproceedings{valassakis2024handdgp,
    author = {Eugene Valassakis and Guillermo Garcia-Hernando},
    title = {{HandDGP}: Camera-Space Hand Mesh Prediction with Differentiable Global Positioning},
    booktitle = {ECCV},
    pages = {479--496},
    year = {2024},
}

@inproceedings{millerdurai2026egoforce,
    author = {Christen Millerdurai and Shaoxiang Wang and Yaxu Xie and Vladislav Golyanik and Didier Stricker and Alain Pagani},
    title = {{EgoForce}: Forearm-Guided Camera-Space {3D} Hand Pose from a Monocular Egocentric Camera},
    booktitle = {SIGGRAPH},
    pages = {45:1--45:12},
    year = {2026},
}
}

\clearpage
\appendix
In this supplementary material, Sec.~\ref{sec:egohandbench_comparison} reports EgoHandBench benchmarks, Sec.~\ref{sec:appendix_mask_comparison} analyzes mask-guided hand localization, and Sec.~\ref{sec:appendix_protocol} defines the evaluation metrics. We then detail the training losses and the binary keyframe pool for sparse bundle adjustment in Sec.~\ref{sec:appendix_training} and Sec.~\ref{sec:binary_keyframe_pool}, and present additional qualitative results in Sec.~\ref{sec:additional_qualitative}.

\section{Fair Comparison on EgoHandBench}
\label{sec:egohandbench_comparison}

EgoHandBench~\citep{egohandbench_supp} standardizes camera calibration and hand annotations across seven subsets, comprising six in-domain splits from ARCTIC~\citep{fan2023arctic}, H2O~\citep{kwon2021h2o}, HOT3D~\citep{banerjee2025hot3d}, OakInk2~\citep{zhan2024oakink2}, and EgoSuite (Pro and Standard)~\citep{egostandard}, alongside Gen-HumanEgo~\citep{genrobot2026genhumanego} for out-of-domain evaluation. Across its 3,617 training and 688 test clips, we evaluate competing methods at native frame rates in camera space and at 30~FPS in world space.

\begin{table}[b]
    \centering
    \vspace{-8pt}
    \caption{\textbf{EgoHandBench camera-space comparison.} FAcc is reported in \%, Abs. denotes Abs-MPJPE, joint errors are measured in mm, EPE in pixels, and Jitter in mm/frame$^2$.}
    \vspace{-4pt}
    \label{tab:egobench_camera}
    \fontsize{7.5}{8.2}\selectfont
    \renewcommand{\arraystretch}{1.1}
    \setlength{\tabcolsep}{1pt}
    \resizebox{\linewidth}{!}{%
    \begin{tabular}{@{}l*{7}{c}@{}}
    \toprule
    Method & FAcc $\uparrow$ & Abs. $\downarrow$ & MPJPE $\downarrow$ & PA $\downarrow$ & EPE$_{\mathrm{2D}}$ $\downarrow$ & Jitter $\downarrow$ & Score $\uparrow$ \\
    \midrule
    InfiniHand & \underline{98.49} & \textbf{37.66} & \underline{19.26} & \underline{7.45} & \textbf{8.18} & \underline{2.07} & \textbf{84.99} \\
    \midrule
    ACE-Ego-Hand~\citep{liu2026ace} & 97.44 & 58.30 & \textbf{16.91} & \textbf{7.17} & 16.77 & \textbf{1.60} & \underline{82.90} \\
    ViDiHand V1~\citep{wang2026vidihand} & \textbf{99.00} & 65.57 & 23.97 & 7.87 & \underline{10.88} & 3.50 & 80.85 \\
    MINT~\citep{zhu2026mint} & 98.30 & 61.41 & 23.78 & 8.57 & 19.54 & 9.11 & 76.46 \\
    WiLoR~\citep{potamias2024wilor} & 92.10 & \underline{51.83} & 22.48 & 8.79 & 13.73 & 12.89 & 75.37 \\
    EgoHandICL~\citep{xie2026egohandicl} & 92.10 & 51.98 & 22.68 & 8.77 & 13.92 & 13.68 & 75.15 \\
    HandFlow~\citep{xu2026handflow} & 92.53 & 56.92 & 24.46 & 9.33 & 16.59 & 15.49 & 72.97 \\
    AnyHand~\citep{si2026anyhand} & 92.10 & 81.00 & 23.14 & 9.04 & 15.08 & 13.05 & 72.91 \\
    PAD-Hand~\citep{ismayilzada2026padhand} & 91.45 & 68.61 & 25.43 & 9.71 & 23.71 & 12.37 & 70.55 \\
    HaMeR~\citep{pavlakos2024hamer} & 89.70 & 79.94 & 24.50 & 9.56 & 20.67 & 16.43 & 68.60 \\
    HandDGP~\citep{valassakis2024handdgp} & 92.12 & 71.11 & 35.89 & 13.45 & 18.34 & 14.47 & 65.80 \\
    Dyn-HaMR~\citep{yu2025dynhamr} & 94.41 & 155.21 & 42.04 & 9.25 & 15.09 & 14.76 & 65.38 \\
    EgoForce~\citep{millerdurai2026egoforce} & 84.56 & 74.42 & 29.38 & 10.83 & 27.72 & 16.52 & 64.09 \\
    InterWild~\citep{moon2023interwild} & 88.11 & 84.31 & 33.14 & 9.74 & 23.33 & 24.65 & 60.89 \\
    WildHands~\citep{prakash2024wildhands} & 88.01 & 93.81 & 38.18 & 11.14 & 28.11 & 16.05 & 60.61 \\
    OmniHands~\citep{lin2024omnihands} & 84.33 & 102.13 & 29.36 & 10.22 & 25.44 & 47.88 & 54.77 \\
    HaWoR~\citep{zhang2025hawor} & 76.47 & 85.36 & 36.85 & 13.61 & 30.58 & 25.02 & 51.13 \\
    Hamba~\citep{dong2024hamba} & 66.31 & 111.30 & 37.59 & 15.01 & 39.49 & 13.71 & 44.89 \\
    \bottomrule
    \end{tabular}}
\end{table}

Tab.~\ref{tab:egobench_camera} reports camera-space results under the benchmark's evaluation protocol (Sec.~\ref{sec:appendix_protocol}), where the overall Score averages normalized per-dataset metrics with equal weighting across the in-domain and out-of-domain splits. Under this protocol, InfiniHand achieves an overall score of 84.99, with abs-MPJPE 37.66 mm and EPE$_\text{2D}$ 8.18 pixels.

\begin{table}[h]
    \centering
    
    \caption{\textbf{EgoHandBench world-space comparison} W and WA are reported in mm, RTE in \%, and Accel in m/s$^2$.}
    \vspace{-4pt}
    \label{tab:egobench_world}
    \scriptsize
    \setlength{\tabcolsep}{2pt}
    \renewcommand{\arraystretch}{1.1}
    \begin{tabular*}{\linewidth}{@{\extracolsep{\fill}}lcccc@{}}
    \toprule
    Method & W-MPJPE$\downarrow$ & WA-MPJPE$\downarrow$ & RTE$\downarrow$ & Accel$\downarrow$ \\
    \midrule
    WiLoR-SLAM~\citep{potamias2024wilor} & \underline{98.97} & 16.72 & 7.03 & 9.94 \\
    ACE-SLAM~\citep{liu2026ace} & 100.31 & \underline{14.82} & \underline{3.94} & \underline{3.15} \\
    HaWoR~\citep{zhang2025hawor} & 121.92 & 27.18 & 12.58 & 23.54 \\
    MINT~\citep{zhu2026mint} & 478.84 & 32.52 & 9.21 & 8.33 \\
    \midrule
    InfiniHand & \textbf{58.43} & \textbf{14.67} & \textbf{3.67} & \textbf{2.63} \\
    \bottomrule
    \end{tabular*}
    \vspace{-8pt}
\end{table}

For world-space evaluation, we evaluate all baselines using checkpoints trained on EgoHandBench, with WiLoR-SLAM and ACE-SLAM relying on DROID-SLAM~\citep{teed2021droidslam} trajectories rescaled via Metric3D v2~\citep{hu2024metric3dv2}. Tab.~\ref{tab:egobench_world} summarizes the quantitative comparison across 688 clips containing jointly valid hand observations. Notably, InfiniHand achieves a W-MPJPE of 58.43 mm, representing a 41.0\% error reduction over the strongest baseline, WiLoR-SLAM. On the Procrustes-aligned metric, it also substantially outperforms the runner-up, HaWoR, lowering WA-MPJPE from 27.18 mm to 14.67 mm.

\begin{figure*}[!t]
    \centering
    \includegraphics[width=\textwidth]{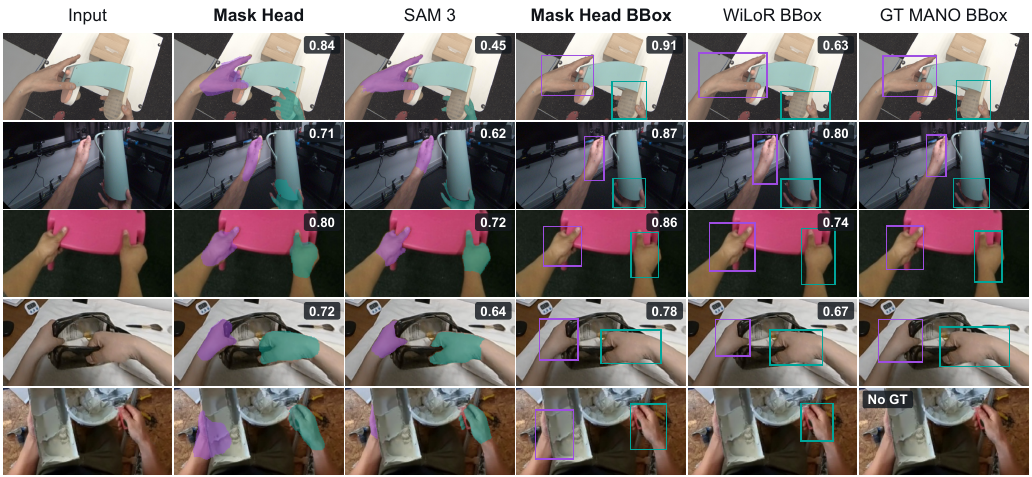}
    \vspace{-16pt}
    \caption{\textbf{Hand localization under occlusion.} For the four annotated examples, rendered ground-truth MANO masks~\citep{romero2017mano} and their tight bounding boxes serve as separate IoU references, whereas the final row displays Ego4D sequences without ground-truth hand annotations.}
    \vspace{-8pt}
    \label{fig:mask_localization_supp}
\end{figure*}

\section{Mask-Guided Hand Localization}
\label{sec:appendix_mask_comparison}
InfiniHand predicts a temporally consistent segmentation mask per hand and extracts its tight enclosing box to crop inputs for downstream hand mesh recovery. Fig.~\ref{fig:mask_localization_supp} compares our masks with single-image SAM~3~\citep{carion2025sam3} predictions and our mask-derived boxes with WiLoR~\citep{potamias2024wilor} detections. In the first example, temporal context preserves clear hand boundaries under motion blur and object interaction, achieving a mask IoU of 0.84 (vs.\ 0.45 for SAM~3) and a box IoU of 0.91 (vs.\ 0.63 for WiLoR), whereas single-image baselines suffer from fragmented masks and truncated boxes. On the unannotated Ego4D~\citep{grauman2022ego4d} sequence, InfiniHand maintains accurate hand crops even under severe hand--object and finger self-occlusion.

\section{Evaluation Metrics}
\label{sec:appendix_protocol}

\subsection{Camera-Space Hand Metrics}

\paragraph{Hand detection.}
A hand is visible if at least one joint has depth $>0.01$~m and projects inside the image. Ground-truth joint boxes are enlarged by $10\%$ and matched to predictions by maximizing $\mathrm{IoU} + 0.5\,\mathbb{I}(\text{same side})$. Matches with identical handedness and positive IoU count as true positives (TP), unmatched visible ground truths count as false negatives (FN), and remaining visible predictions count as false positives (FP). Over $T$ frames,
\begin{equation}
\begin{aligned}
\mathrm{F1}&=\frac{2\mathrm{TP}}{2\mathrm{TP}+\mathrm{FP}+\mathrm{FN}},\\
\mathrm{FAcc}&=\frac{1}{T}\sum_{t=1}^{T}\mathbb{I}[\mathrm{FP}_t=0\ \land\ \mathrm{FN}_t=0],
\end{aligned}
\end{equation}
where F1 aggregates counts across all frames and hand sides, and both metrics lie in $[0,1]$.

\paragraph{Joint errors.}
For a hand-frame instance $i$ comprising $J=21$ joints and wrist $j=0$, let $\hat{\mathbf J}_{i,j}^{\mathrm c}$ and $\mathbf J_{i,j}^{\mathrm c}$ denote the predicted and ground-truth joint positions in meters, respectively. The evaluated prediction $\widetilde{\mathbf J}_{i,j}^{\mathrm c}$ corresponds to the matched prediction, or defaults to the same-side mean MANO hand template positioned along the image-center ray at a depth of $0.5$ m if a visible hand is missed. Measured in millimeters, the absolute, wrist-relative, and Procrustes-aligned (PA) joint errors are defined as
\begin{equation}
\begin{aligned}
e_{\mathrm{Abs}}(i)
&=\frac{10^3}{J}\sum_j
\left\|\widetilde{\mathbf J}_{i,j}^{\mathrm c}-\mathbf J_{i,j}^{\mathrm c}\right\|_2,\\
e_{\mathrm{MP}}(i)
&=\frac{10^3}{J}\sum_j
\left\|\widetilde{\mathbf J}_{i,j,\mathrm{rel}}^{\mathrm c}
-\mathbf J_{i,j,\mathrm{rel}}^{\mathrm c}\right\|_2,\\
e_{\mathrm{PA}}(i)
&=\frac{10^3}{J}\sum_j
\left\|A_i^*\!\left(\hat{\mathbf J}_{i,j}^{\mathrm c}\right)
-\mathbf J_{i,j}^{\mathrm c}\right\|_2,
\end{aligned}
\end{equation}
where each wrist-relative joint position is centered by subtracting its respective wrist joint, and $A_i^*$ represents a similarity transformation fitted to the matched hand instance. For missed hands, PA error is computed as the root-mean-square (RMS) distance of the ground-truth joints from their centroid.

\paragraph{2D reprojection error.}
For the set of visible ground-truth joints $\mathcal U$, let $d_{i,j}$ denote the pixel distance between the calibrated projections of $\widetilde{\mathbf J}_{i,j}^{\mathrm c}$ and $\mathbf J_{i,j}^{\mathrm c}$. Given the image diagonal $D_i=\sqrt{W_i^2+H_i^2}$ in pixels,
\begin{equation}
\mathrm{EPE}_{\mathrm{2D}}=\frac{1}{|\mathcal U|}\sum_{(i,j)\in\mathcal U}
\min\!\left(d_{i,j},D_i\right).
\end{equation}

\paragraph{Jitter.}
Let $\mathcal V_{\mathrm{jit}}$ denote the set of consecutive true-positive frame triplets for the same hand $h$. Temporal jitter is computed by averaging the norm of the second-order finite difference of the predicted joint positions:
\begin{equation}
\mathrm{Jitter}=10^3
\underset{(t,h,j)\in\mathcal V_{\mathrm{jit}}}{\operatorname{mean}}
\left\|
\hat{\mathbf J}_{t+1,j}^{\mathrm c,h}
-2\hat{\mathbf J}_{t,j}^{\mathrm c,h}
+\hat{\mathbf J}_{t-1,j}^{\mathrm c,h}
\right\|_2.
\end{equation}
Jitter is reported in units of mm/frame$^2$ without normalization by elapsed time. Finally, 3D joint errors are pooled at the hand-frame level, EPE$_{\mathrm{2D}}$ at the visible-joint level, and jitter at the joint-triplet level.

\subsection{World-Space Hand Metrics}

\paragraph{W-MPJPE and WA-MPJPE.}
Predicted and ground-truth joints are expressed in their respective camera frames at the first frame with valid ground-truth extrinsics, with mutually valid observations partitioned into continuous runs per hand. Both metrics evaluate non-overlapping windows $\mathcal W$ of 2--100 frames within each run. Given a Sim(3) alignment $\mathcal A_{\mathcal S}$ fitted from predicted to ground-truth joints over frames $\mathcal S$, the window error in millimeters (for coordinates in meters) is
\begin{equation}
E(\mathcal W;\mathcal S)=\frac{10^3}{J|\mathcal W|}
\sum_{t\in\mathcal W}\sum_j
\|\mathcal A_{\mathcal S}(\hat{\mathbf J}_{t,j}^{\mathrm w,s})-\mathbf J_{t,j}^{\mathrm w,s}\|_2.
\end{equation}
W-MPJPE fits $\mathcal A_{\mathcal S}$ on the first two frames of $\mathcal W$, whereas WA-MPJPE fits it over the entire window. Window errors are frame-count weighted when aggregated across hands and clips.

\paragraph{Root trajectory error.}
Over each continuous run $\mathcal R$, Root Trajectory Error (RTE) rigidly aligns the predicted wrist trajectory at fixed scale via $\mathcal G_{\mathcal R}$. With ground-truth path length $L_{\mathcal R}$, the normalized residual at frame $t$ is
\begin{equation}
\begin{aligned}
L_{\mathcal R}
&=\sum_{t\in\mathcal R\setminus\{\min\mathcal R\}}
\left\|\mathbf J_{t,0}^{\mathrm w,s}-\mathbf J_{t-1,0}^{\mathrm w,s}\right\|_2,\\
e_{\mathrm{RTE}}(t)
&=100\,
\frac{\left\|\mathcal G_{\mathcal R}\!\left(\hat{\mathbf J}_{t,0}^{\mathrm w,s}\right)-\mathbf J_{t,0}^{\mathrm w,s}\right\|_2}
{L_{\mathcal R}}.
\end{aligned}
\end{equation}
RTE averages $e_{\mathrm{RTE}}(t)$ as a percentage, excluding zero-length trajectories.

\paragraph{Acceleration error.}
At 30~FPS ($\Delta=1/30$~s), we define finite-difference acceleration as
\begin{equation}
\mathbf a_t(\mathbf x)=
\frac{\mathbf x_{t+1}-2\mathbf x_t+\mathbf x_{t-1}}{\Delta^2}.
\end{equation}
Averaging across valid interior samples $\mathcal V_{\mathrm{acc}}$ (excluding the first and last frames of each continuous run) yields Acceleration Error (Accel) in m/s$^2$,
\begin{equation}
\mathrm{Accel}=
\underset{(t,s,j)\in\mathcal V_{\mathrm{acc}}}{\operatorname{mean}}
\left\|
\mathbf a_t\!\left(\hat{\mathbf J}_{\cdot,j}^{\mathrm w,s}\right)
-\mathbf a_t\!\left(\mathbf J_{\cdot,j}^{\mathrm w,s}\right)
\right\|_2.
\end{equation}

\section{Loss Functions}
\label{sec:appendix_training}
All loss terms are averaged over valid annotations, with empty valid sets contributing zero to the objective. For clarity, outer sample-averaging operators are omitted from the definitions of individual mask, joint, and MANO error terms, and hat accents denote predicted quantities.

\subsection{Mask Losses}
For scalar pixel entries $m_p$ and $\hat m_p$ of a valid ground-truth and predicted mask pair $(\mathbf m,\hat{\mathbf m})$ containing $N_{\mathrm{pix}}$ pixels, the binary cross-entropy and soft Dice losses are defined as
\begin{equation}
\begin{aligned}
\ell_{\mathrm{BCE}}&=-\frac{1}{N_{\mathrm{pix}}}\sum_{p=1}^{N_{\mathrm{pix}}}\left[m_p\log\hat m_p+(1-m_p)\log(1-\hat m_p)\right],\\
\ell_{\mathrm{Dice}}&=1-\frac{2\sum_p\hat m_p m_p+1}{\sum_p\hat m_p+\sum_p m_p+1}.
\end{aligned}
\end{equation}
For numerical stability, BCE is computed directly from raw logits, whereas the soft Dice loss operates on sigmoid-activated probabilities. Averaging each term across all valid masks yields $\mathcal L_{\mathrm{mask}}=\lambda_{\mathrm{BCE}}\mathcal L_{\mathrm{BCE}}+\lambda_{\mathrm{Dice}}\mathcal L_{\mathrm{Dice}}$.

\subsection{MANO Reconstruction Losses}
The 2D joint predictions from the MANO regression head are supervised directly in normalized crop coordinates, with the valid annotation set excluding missing, non-finite, or out-of-bounds ground-truth joints.

Let $\operatorname{Rot}(\cdot)$ map an axis-angle vector to its corresponding rotation matrix. For rotations $\hat{\mathbf R}$ and $\mathbf R$, the geodesic distance is
\begin{equation}
c=\tfrac{\operatorname{tr}(\hat{\mathbf R}^{\top}\mathbf R)-1}{2},\quad
d_{\mathrm{SO}(3)}=\arccos(c_\epsilon),
\end{equation}
where $c_\epsilon$ is $c$ clipped to $[-1+\epsilon,1-\epsilon]$.
Both global orientation and local hand pose are supervised using this geodesic distance, whereas shape parameters and camera-frame MANO translation are supervised via coordinate-wise squared error:
\begin{equation}
\begin{aligned}
\mathcal L_{\mathrm{orient}}&= d_{\mathrm{SO}(3)}\!\left(\operatorname{Rot}(\hat{\boldsymbol\Phi}^{\mathrm c}),\operatorname{Rot}(\boldsymbol\Phi^{\mathrm c})\right),\\
\mathcal L_{\mathrm{pose}}&=\frac{1}{15}\sum_{k=1}^{15}d_{\mathrm{SO}(3)}\!\left(\operatorname{Rot}(\hat{\boldsymbol\Theta}_k),\operatorname{Rot}(\boldsymbol\Theta_k)\right),\\
\mathcal L_{\mathrm{shape}}&=\frac{1}{10}\|\hat{\boldsymbol\beta}-\boldsymbol\beta\|_2^2,\ 
\mathcal L_{\mathrm{trans}}=\frac{1}{3}\|\hat{\mathbf t}^{\mathrm c}-\mathbf t^{\mathrm c}\|_2^2.
\end{aligned}
\end{equation}
The 2D joint, 3D joint, and reprojection loss terms are defined as
\begin{equation}
\begin{aligned}
\mathcal L_{\mathrm{2D}}&=\|\hat{\mathbf p}-\mathbf p\|_1,\\
\mathcal L_{\mathrm{joints}}^{\mathrm c}&=\frac{1}{3}\|\hat{\mathbf J}^{\mathrm c}-\mathbf J^{\mathrm c}\|_1,\\
\mathcal L_{\mathrm{reproj}}&=\|\bar\pi_{\mathbf K}(\hat{\mathbf J}^{\mathrm c})-\mathbf p^{\mathrm{img}}\|_1.
\end{aligned}
\end{equation}
Here, $\bar\pi_{\mathbf K}$ projects points into normalized image coordinates, and $\mathbf p^{\mathrm{img}}$ denotes the corresponding targets. Validity masks are applied on a per-joint basis, with both projections and ground-truth targets sharing identical camera intrinsics and image normalization. Unlike $\mathcal L_{\mathrm{2D}}$, the reprojection term directly supervises the joints decoded from the MANO layer.

\subsection{World-Space Losses}
\paragraph{World-space joints.}
After applying the anchor transformation from Equation (9) of the main paper alongside identical sample-level normalization to both predictions and ground-truth targets, the world-space joint loss compares the predicted joints against the dataset-provided annotations using a robust element-wise smooth-$L_1$ penalty $\rho_\epsilon$ (summed across spatial coordinates):
\begin{equation}
\mathcal L_{\mathrm{joints}}^{\mathrm{w}}
=
\frac{1}{N_{\mathrm{J}}}
\sum_{t,s,j}
v_{t,s}^{\mathrm{J}}
\rho_\epsilon
\left(
\widetilde{\hat{\mathbf J}}_{t,j}^{\mathrm{w},s}
-
\widetilde{\mathbf J}_{t,j}^{\mathrm{w},s}
\right).
\end{equation}
Here, $\rho_\epsilon(r)=r^2/(2\epsilon)$ for $|r|<\epsilon$ and $|r|-\epsilon/2$ otherwise, while $v_{t,s}^{\mathrm{J}}$ indicates hand validity and $N_{\mathrm{J}}$ denotes the total number of valid joint coordinates. The supervision target comes directly from the original dataset joint annotations.

\paragraph{Camera supervision.}
The camera supervision loss combines absolute and relative pose terms as $\mathcal L_{\mathrm{cam}}=\alpha_{\mathrm{abs}}\mathcal L_{\mathrm{abs\text{-}pose}}+\alpha_{\mathrm{rel}}\mathcal L_{\mathrm{rel\text{-}pose}}$. Specifically, the absolute-pose term supervises the anchor-normalized camera translation, rotation quaternion, and field of view. Since $\mathbf q$ and $-\mathbf q$ represent the same rotation, we set the sign-aligned target $\mathbf q_t^\star$ to $\mathbf q_t$ when $\langle\hat{\mathbf q}_t,\mathbf q_t\rangle\geq 0$ and to $-\mathbf q_t$ otherwise.
With $\rho_\epsilon$ summed across coordinates, the framewise translation, quaternion, and field-of-view costs and their aggregate are
\begin{equation}
\begin{aligned}
\ell_{\mathrm u,t}&=\rho_\epsilon(\widetilde{\hat{\mathbf u}}_t-\widetilde{\mathbf u}_t),\\
\ell_{\mathrm q,t}&=\rho_\epsilon(\hat{\mathbf q}_t-\mathbf q_t^\star),\\
\ell_{\mathrm{fov},t}&=\rho_\epsilon(\hat{\boldsymbol{\gamma}}_t-\boldsymbol{\gamma}_t),\\
\mathcal L_{\mathrm{abs\text{-}pose}}
&=\frac{\sum_{t\in\mathcal V_{\mathrm{cam}}}
\bigl(\ell_{\mathrm u,t}+\ell_{\mathrm q,t}+\ell_{\mathrm{fov},t}\bigr)}
{9|\mathcal V_{\mathrm{cam}}|}.
\end{aligned}
\end{equation}
Here, $\mathcal V_{\mathrm{cam}}$ denotes the set of frames with valid camera annotations, $\mathbf q_t$ represents a unit quaternion, and $\boldsymbol{\gamma}_t$ specifies the horizontal and vertical field-of-view angles.

To regularize local camera motion independently of the anchor choice, let $\mathcal P_{\mathrm{cam}}$ be the ordered pairs $(i,j)$ of distinct frames in the pose-reference window $\mathcal W_{\mathrm{pose}}$. Let $\mathbf C_t=\left[\begin{smallmatrix}\mathbf R_t&\mathbf u_t\\\mathbf 0^\top&1\end{smallmatrix}\right]$ denote the homogeneous camera-to-world transform of $P_t$. The relative transform from frame $i$ to frame $j$ is $\mathbf C_{j\leftarrow i}=\mathbf C_j^{-1}\mathbf C_i$. The rotation and translation costs give the relative-pose loss as
\begin{equation}
\begin{aligned}
\ell_{\mathrm R,ji}&=d_{\mathrm{SO}(3)}(\Delta\hat{\mathbf R}_{ji},\Delta\mathbf R_{ji}),\\ 
\ell_{\mathrm u,ji}&=\lambda_{\mathrm{rel},\mathrm u}\|\Delta\hat{\mathbf u}_{ji}-\Delta\mathbf u_{ji}\|_1,\\
\mathcal L_{\mathrm{rel\text{-}pose}}
&=\frac{1}{N_{\mathrm{pair}}}
\sum_{(i,j)\in\mathcal P_{\mathrm{cam}}}
v_i^{\mathrm{cam}}v_j^{\mathrm{cam}}
\bigl(\ell_{\mathrm R,ji}+\ell_{\mathrm u,ji}\bigr).
\end{aligned}
\end{equation}
Here, $v_t^{\mathrm{cam}}$ indicates camera annotation validity, $N_{\mathrm{pair}}$ counts the valid ordered frame pairs, and $\Delta\mathbf R_{ji}$ and $\Delta\mathbf u_{ji}$ denote the rotation and translation components of $\mathbf C_{j\leftarrow i}$, respectively. Note that the camera translations entering this term have already been normalized by $\kappa$ and are therefore not rescaled by this factor again.

\paragraph{Hand temporal consistency.}
The hand regularization terms in $\mathcal L_{\mathrm{temp}}$ constrain temporal smoothness of translation and consistency of MANO shape parameters. For each triplet of consecutive frames containing valid predictions for the same hand, translations are expressed in the original camera coordinate axes, and the smoothness loss is computed as the $L_1$ norm of the velocity change scaled by $2(0.1\,\mathrm{s})^2/(\Delta_-+\Delta_+)$, where $\Delta_-$ and $\Delta_+$ denote the two consecutive frame intervals. Meanwhile, the shape consistency term penalizes the mean squared deviation of each valid shape vector from the hand's clip-level mean shape, with gradient backpropagation stopped through the mean.

\section{Binary Keyframe Pooling for BA}
\label{sec:binary_keyframe_pool}
We maintain a chronological keyframe pool $\mathcal K_{\mathrm{BA}}=\{(\xi_i,b_i)\}_{i=1}^{n}$, where $\xi_i$ stores a frame together with its geometric state and $b_i\in\{0,1\}$ indicates whether it has survived a prior refinement pass. Newly inserted entries are initialized with $b_i=0$. Whenever $|\mathcal K_{\mathrm{BA}}|>N$, we subsample every $K$-th entry starting from the oldest element and jointly optimize the selected camera poses using the DROID-SLAM backend~\citep{teed2021droidslam}. The pool is subsequently updated according to
\begin{equation}
\begin{aligned}
\mathcal S&=\{\xi_{1+mK}\mid m\geq0,\ 1+mK\leq |\mathcal K_{\mathrm{BA}}|\},\\
b_i^{+}&=b_i\oplus\mathbb{I}[\xi_i\in\mathcal S],\\
\mathcal K_{\mathrm{BA}}^{+}&=\{(\xi_i,b_i^{+})\mid(\xi_i,b_i)\in\mathcal K_{\mathrm{BA}},\ b_i^{+}=1\},
\end{aligned}
\label{eq:keyframe_pool}
\end{equation}
All selected frames undergo pose refinement prior to the pool update. Upon first selection, a frame is retained in the pool $(0\!\to\!1)$, whereas a subsequent selection retires it $(1\!\to\!0)$. Meanwhile, unselected previously retained frames remain in the pool while unselected new frames are discarded, yielding dense local constraints alongside sparse historical constraints. The threshold $N$ sets the refinement trigger, and $K$ controls its temporal sparsity. Note that bundle adjustment (BA) updates solely the camera poses for world-space transformation, while hand mesh reconstruction remains purely feed-forward.

\begin{figure*}[p]
    \centering
    \includegraphics[width=\textwidth,height=.90\textheight,keepaspectratio]{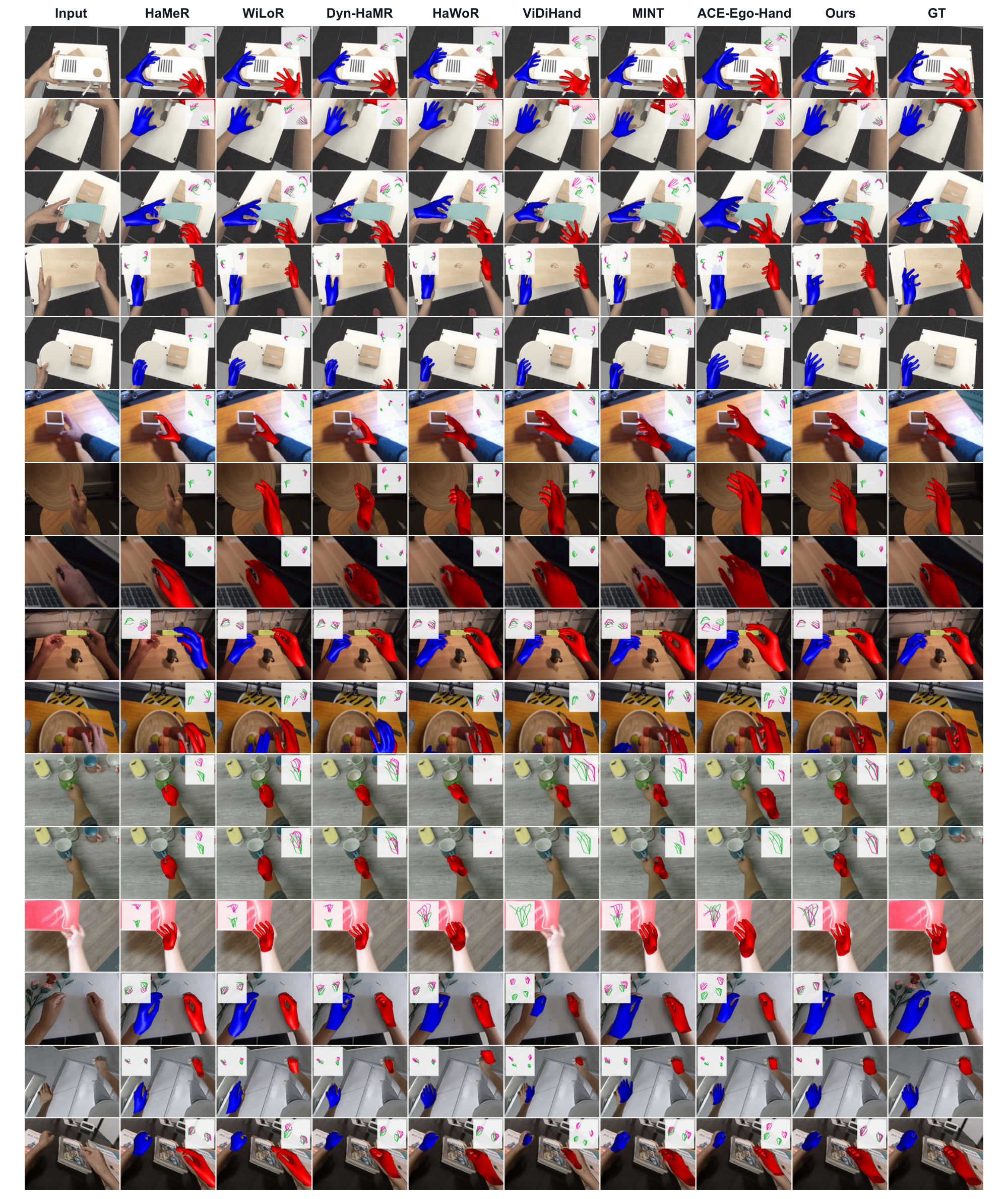}
    \caption{\textbf{Camera-space hand reconstruction.} Each row displays the input RGB frame, hand mesh projections from the compared methods, and the ground truth. Top-down insets compare ground-truth joints (green) with predicted joints (magenta) in the camera $X$--$Z$ plane, complementing the image overlays with a view of depth alignment. Examples are drawn from ARCTIC~\citep{fan2023arctic}, HOT3D~\citep{banerjee2025hot3d}, HOI4D~\citep{liu2022hoi4d}, and Gen-HumanEgo~\citep{genrobot2026genhumanego}.}
    \label{fig:camera_qualitative_supp}
\end{figure*}

\begin{figure*}[p]
    \centering
    \includegraphics[width=\textwidth,height=.90\textheight,keepaspectratio]{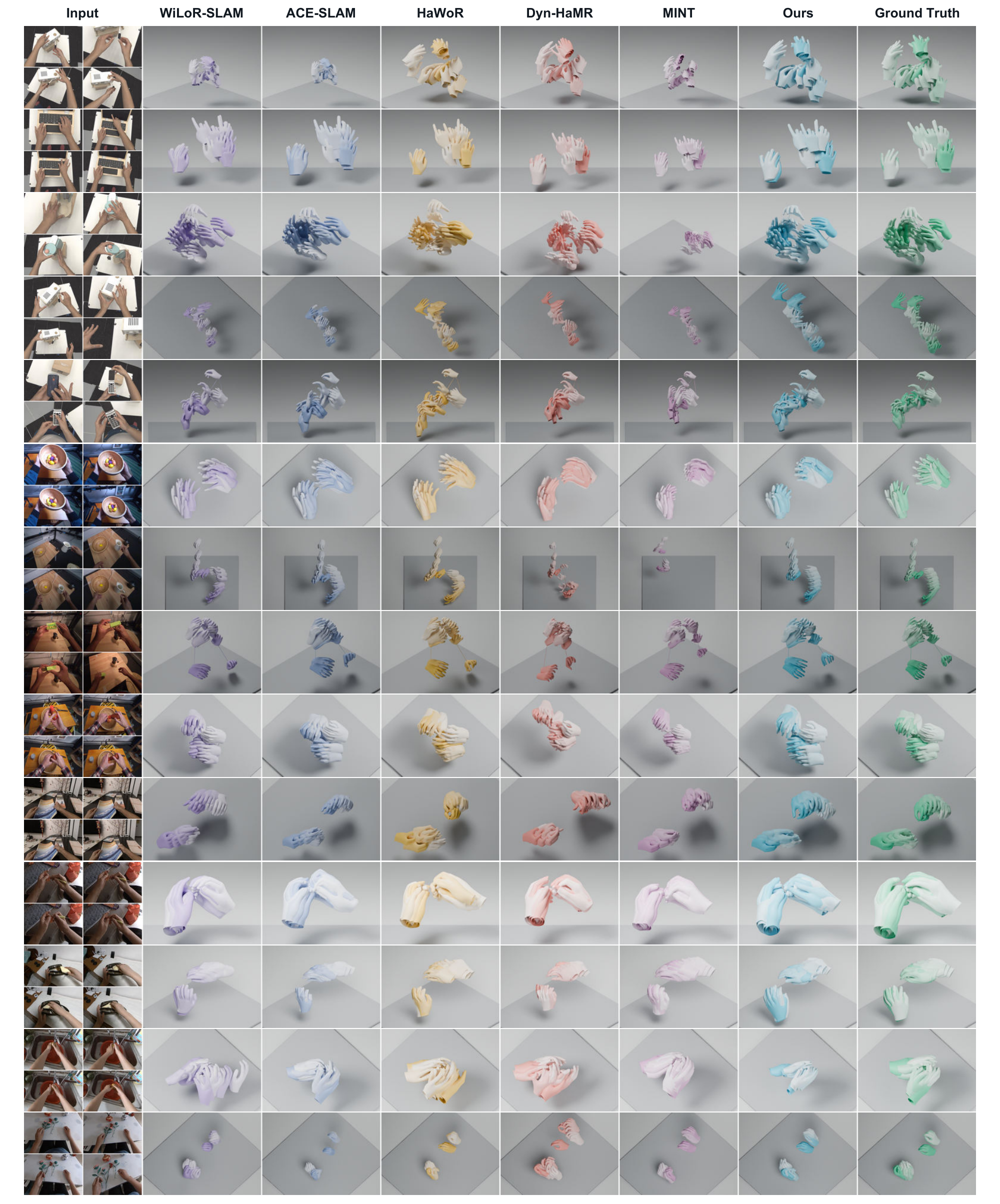}
    \caption{\textbf{World-space hand motion reconstruction.} The first column displays four sampled input frames per sequence, while the remaining columns render hand meshes across multiple time steps within a unified world coordinate system. These visualizations highlight global hand trajectory accuracy and bimanual spatial consistency relative to the ground truth.}
    \label{fig:world_qualitative_supp}
\end{figure*}

\section{Additional Qualitative Results}
\label{sec:additional_qualitative}
Fig.~\ref{fig:camera_qualitative_supp} showcases additional camera-space qualitative results, where 2D mesh overlays illustrate image alignment and fine finger articulation, and the $X$--$Z$ bird's-eye insets highlight relative depth accuracy. Even under heavy hand-object contact and self-occlusion, InfiniHand avoids the depth errors and distorted finger poses observed in competing baselines. Fig.~\ref{fig:world_qualitative_supp} further illustrates reconstructed world-space trajectories across both in-domain and out-of-domain sequences. Across both evaluation regimes, InfiniHand suppresses trajectory drift and temporal jitter, aligning most closely with the ground truth and demonstrating accurate, temporally consistent world-space hand reconstruction.

\clearpage

\end{document}